\documentclass[11pt]{article}

\usepackage[final]{acl}

\usepackage{times}
\usepackage{latexsym}
\usepackage{amsmath}
\usepackage{amssymb}
\usepackage{booktabs}
\usepackage{graphicx}
\usepackage{enumitem}
\usepackage{multirow}
\usepackage[table]{xcolor} 
\usepackage{xurl}
\usepackage{tabularx}
\usepackage{array}
\usepackage{ragged2e}
\usepackage{threeparttable}
\usepackage{makecell}
\usepackage{geometry}
\usepackage{booktabs}
\usepackage{tabularx}
\usepackage{xcolor}
\usepackage{colortbl}
\usepackage{ragged2e}
\usepackage{hyperref}
\usepackage{subcaption} 
\usepackage[normalem]{ulem}

\usepackage[most]{tcolorbox}
\usepackage{enumitem}

\usepackage[T1]{fontenc}

\usepackage[utf8]{inputenc}

\usepackage{microtype}

\usepackage{inconsolata}

\usepackage{graphicx}

\title{
SAEExplainer: Interpreting SAE Features with Activation-Guided Preference Optimization
}

\author{
\textbf{Jingyi He}\textsuperscript{1,2},
\textbf{Haiyan Zhao}\textsuperscript{3},
\textbf{Ruxue Shi}\textsuperscript{4},
\textbf{Yanguang Liu}\textsuperscript{3},
\textbf{Xin Wang}\textsuperscript{4},
\textbf{Fei Sun}\textsuperscript{5},\\
\textbf{Mengnan Du}\textsuperscript{1,\textdagger}\\
\textsuperscript{1}The Chinese University of Hong Kong, Shenzhen \,
\textsuperscript{2}Shanghai Jiao Tong University \,\\
\textsuperscript{3}NJIT \, 
\textsuperscript{4}Jilin University \,
\textsuperscript{5}Institute of Computing Technology, CAS\\
\texttt{sunrain-H@sjtu.edu.cn, mengnandu@cuhk.edu.cn}\\
\small\textsuperscript{\textdagger}Corresponding author.
}

\begin{document}
\maketitle
\begin{abstract}
Although Sparse Autoencoders (SAEs) have mitigated the opacity of large language models (LLMs) by decomposing dense representations into sparse features, explaining these features still remains a central challenge. Current explanation methods, however, typically operate within an open-loop paradigm, failing to leverage mechanistic feedback for further refinement. In this paper, we propose \textbf{SAEExplainer}, a training framework that utilizes activation scores as an objective reward signal to train the model for self-correction and iterative bootstrapping. By iteratively verifying and correcting foundational explanations through a two-round optimization process, SAEExplainer achieves continuous improvement in its explanatory capabilities. This mechanism significantly reduces explanation hallucinations and reinforces causal triggering patterns. Extensive experiments demonstrate our approach improves upon established baselines across most metrics, especially in causal triggering and discriminative activation. 
The code is available at \url{https://github.com/he-jingyi/SAEExplainer}.
\end{abstract}

\section{Introduction}
Recently, Sparse Autoencoders (SAEs) have attracted significant attention within the Mechanistic Interpretability community. On the one hand, researchers increasingly utilize SAEs to deconstruct and explain the internal behaviors of LLMs~\cite{ferrando2025know}. On the other hand, the isolated SAE features have proven to be highly effective in steering model behaviors~\cite{he2025sae,li2025feature}. Nevertheless, while these features successfully disentangle dense activations, they remain continuous, high-dimensional vectors that demand further semantic grounding. Explaining these SAE features into accurate natural languages remains a critical and highly challenging task.

Currently, methodologies for interpreting SAE feature vectors primarily fall into two categories. The first approach prompts an external LLM to summarize highly activating texts into natural language explanations \citep{bills2023language,paulo2024automatically}. The second approach pairs these generated explanations with SAE feature vectors to train a model that directly verbalizes their semantics \citep{karvonen2025activation,pan2026latentqa}.
Although existing research has achieved preliminary results, significant challenges remain.
First, while focusing on generation quality, existing methods rarely exploit explanations that appear plausible but fail to trigger the target features. This absence prevents models from distinguishing linguistically fluent from mechanistically faithful explanations, significantly increasing the risks of explanation hallucinations and over-generalization. Second, current approaches rely on an open-loop, single-pass paradigm. Although evaluation mechanisms exist, their feedback is used merely for scoring or filtering rather than as learning signals to optimize the explainer. Consequently, continuous improvement of explanatory capabilities is hindered.

Motivated by these limitations, we ask: \uline{Can we establish a closed-loop mechanism that leverages mechanistic feedback to verify and correct explanations?} To address this, we propose \textbf{SAEExplainer}, a novel framework that utilizes activation scores as an objective reward signal to train the model for self-correction and continuous improvement.
We argue that a faithful explanation should reliably trigger the target feature in generated texts; consistent failure implies it is overly broad, incomplete, or hallucinated.
Building on this core intuition, we operationalize the target model's objective activation scores as a verifiable reward. Specifically, after equipping the model with basic explanatory capabilities, we prompt the model to generate multiple candidate explanations per feature and filter them using real activation feedback. High-quality explanations verified at the activation level are selected as positive samples, while we deliberately retain plausible-sounding but non-activating explanations as hard negatives. By constructing these preference pairs and applying Direct Preference Optimization (DPO), the model is forced to sharpen its discriminative ability via contrastive learning, empowering it to self-correct and refine its outputs.
Crucially, this mechanism drives the progressive enhancement of its explanatory capabilities. As the model's proficiency grows, it yields more precise candidate explanations, which in turn construct even higher-quality preference datasets for the subsequent training round. Through this, the model achieves sustained self-verification and correction, effectively reducing explanatory hallucinations and ensuring that the final natural language explanations strictly align with the true mechanistic activation conditions. 
Extensive experiments demonstrate that SAEExplainer yields highly competitive performance across metrics, achieving substantial gains over baselines in generating causally faithful explanations.
Our main contributions are summarized as follows:
\begin{itemize}[leftmargin=10pt, topsep=-2pt, itemsep=1pt, partopsep=1pt, parsep=1pt]
    \item We propose SAEExplainer, a novel training framework that equips the model with the capability to comprehend feature vectors and translate them into natural language explanations.
    \item SAEExplainer leverages objective target model activations to drive progressive enhancement, bypassing the constraints of single-pass generation and mitigating explanatory hallucinations.
    \item Extensive experiments across diverse LLMs demonstrate that SAEExplainer generates explanations with superior faithfulness and stronger causal validity.
\end{itemize}

\section{Preliminary}

\subsection{Sparse Autoencoders}

SAEs \citep{gao2025scaling,bricken2023monosemanticity} are an unsupervised neural network architecture consisting of an encoder and a decoder. Taking a dense activation vector $x \in \mathbb{R}^{d_{\text{model}}}$ at a specific layer and token position within the LLM as input, the encoder applies a non-linear activation function to map it into high-dimensional sparse activation vector $f \in \mathbb{R}^{d_{\text{sae}}}$, satisfying $d_{\text{sae}} \gg d_{\text{model}}$. Driven by a reconstruction loss from the decoder and an introduced sparsity penalty, the SAE functions as an effective disentangler, transforming the dense, polysemantic internal activations of the LLM into sparse and monosemantic feature representations.

Formally, the SAE encoder extracts the sparse activation vector $f$ and the decoder attempts to reconstruct the original activation signal $\hat{x}$ using the activation vector $f$:
\begin{equation}
    f = \sigma(W_e(x - b_{\text{pre}}) + b_e),
\end{equation}
\begin{equation}
    \hat{x} = W_d f + b_{\text{dec}},
\end{equation}
where $W_e \in \mathbb{R}^{d_{\text{sae}} \times d_{\text{model}}}$ and $W_d \in \mathbb{R}^{d_{\text{model}} \times d_{\text{sae}}}$ are the encoder and decoder weight matrices respectively, while $b_{\text{pre}}$, $b_e$ and $b_{\text{dec}}$ are biases, and $\sigma$ is a non-linear activation.
The model is trained to balance reconstruction fidelity and feature sparsity by minimizing the following loss function:
\begin{equation}
    \mathcal{L} = \underbrace{\|x - \hat{x}\|_2^2}_{\text{Reconstruction Loss}} + \underbrace{\lambda \|f\|_1}_{\text{Sparsity Penalty}}.
\end{equation}

\begin{figure*}[t]
    \centering
    \includegraphics[width=\textwidth]{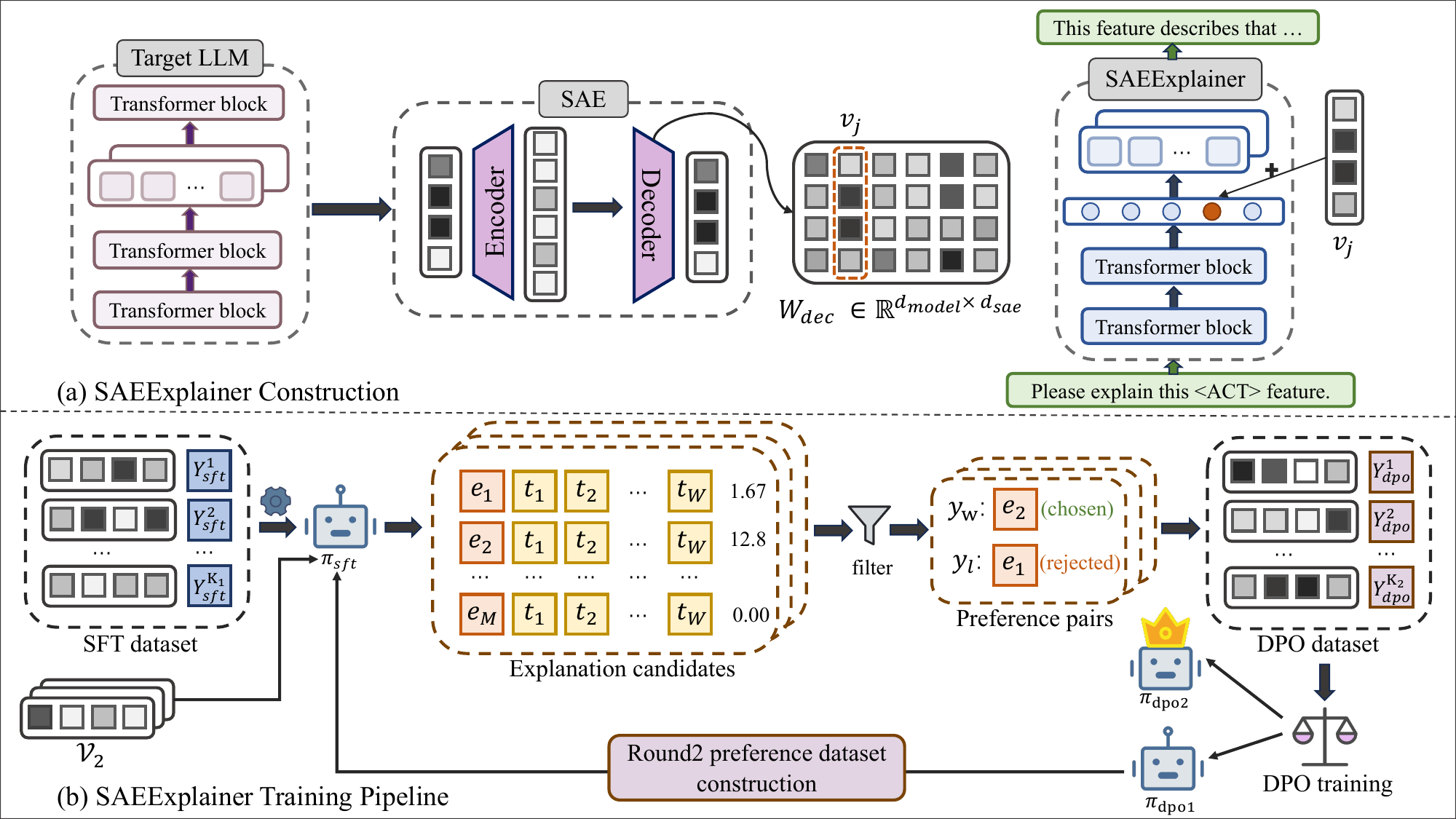}
    \caption{Overview of the SAEExplainer framework, which directly maps SAE feature vectors to natural-language explanations through an explainer language model. The training pipeline first initializes the explainer with SFT, then constructs mechanistic preference pairs by scoring candidate explanations with target-model SAE activations, and finally applies iterative DPO to improve explanation faithfulness and reduce hallucinations.}
    \label{fig:figure1}
\end{figure*}

\subsection{SAE Feature Explanation}
While the emergence of SAEs has greatly propelled our understanding of LLM inner mechanisms, the resulting SAE feature vectors still requires post-hoc semantic interpretation. Formally, for an SAE with a feature dimension $d_{\text{sae}}$, the decoding reconstruction is expressed as a linear combination of column vectors:
\begin{equation}
    \hat{x} \approx \sum_{i=1}^{d_{\text{sae}}} f_i v_i + b_{\text{dec}},
\end{equation}
where $f_i$ is a dynamic scalar in the sparse activation vector $f \in \mathbb{R}^{d_{\text{sae}}}$ that records the feature's local activation intensity in a given context, whereas $v_i \in \mathbb{R}^{d_{\text{model}}}$ (the $i$-th column of the decoder weight matrix $W_d$) is a high-dimensional dense vector determining the concept's global semantic direction in the latent space. Thus, the feature explanation task aims to generate a natural language description $e_i$ that precisely captures the semantic and structural essence of the feature vector $v_i$.

Current mainstream methods, however, operate as open-loop systems lacking a scalable self-improvement mechanism, which renders the generated explanations prone to hallucination. To overcome this limitation, we aim to train an Explainer LLM that goes beyond foundational explanation generation by leveraging objective activation scores for self-correction and bootstrapping.

\section{Methodology}

In this section, we present \textbf{SAEExplainer}, designed to address the issues of overly broad explanations and hallucinations inherent in open-loop explanation generation systems. Our overall pipeline is illustrated in Figure \ref{fig:figure1}. SAEExplainer constructs a preference dataset leveraging the authentic activation feedback from the target model, enabling contrastive learning to achieve explanation correction and refinement.
\subsection{SAE Explainer Construction}
We apply the SAEExplainer framework to a base LLM, and train the model to generate corresponding explanations for the input feature vectors. Because directly feeding the feature vector as input would lead to modality mismatch and early attention disruption, we opt for a residual stream injection mechanism to enable the model to genuinely comprehend the intrinsic semantics of the feature vector. Specifically, we construct the input $X^i$ for a given SAE feature vector $v_i$, which is formulated as follows:
\begin{equation}
    X^i = (x_\textrm{text}, v_i),
\end{equation}
where $x_\textrm{text}$ is a fixed instruction template (Appendix \ref{prompt sft}) containing a designated placeholder token $\langle \mathrm{ACT} \rangle$ (e.g., \textit{``The internal neural feature represented by $\langle \mathrm{ACT} \rangle$ is provided...''}). Following the setting in Activation Oracles \citep{karvonen2025activation}, we locate the sequence position $p$ of the $\langle \mathrm{ACT} \rangle$ placeholder and inject the vector $v_i$ after the second Transformer layer (Layer 1) during the forward pass. We then modify the residual stream activation at this specific position by adding a norm-matched steering vector:
\begin{equation}
    h'_p = h_p + \frac{\|h_p\|}{\|v_i\|} \cdot v_i,
\end{equation}
where $h_p$ is the original activation at position $p$ in layer 1, and $\|\cdot\|$ denotes the $\ell_2$ norm. 
This mechanism scales the injected vector to match the residual stream, yielding a modified hidden state $h'_p$ that propagates to decode the explanation.

\subsection{Initialize Explainer with SFT}
\label{sec:sft_warmup}

To mitigate the cold-start problem where an untrained explainer struggles to comprehend the SAE feature vectors, we introduce a SFT warm-up stage. Assuming the SAE applied to the $l$-th layer of the target LLM encompasses a total of $N$ features, we randomly sample a subset of $K_1$ feature vectors, denoted as $\mathcal{V}_1 = \{v_i\}_{i=1}^{K_1}$. We utilize their corresponding Neuronpedia explanations \citep{neuronpedia} as ground-truth labels $Y_\textrm{sft}$. Using the input formulation $X^i$, the explainer is fine-tuned on the dataset $\mathcal{D}_\textrm{sft} = \{(X^i, Y_\textrm{sft}^i)\}_{i=1}^{K_1}$. This brief warm-up stage establishes the model's fundamental instruction-following format and basic feature-decoding capabilities prior to the preference alignment.

\subsection{Mechanistic Preference Construction}
\label{sec:preference_construction}

The SFT stage yields a base explainer $\pi_{\text{sft}}$. To construct a preference dataset for DPO training, we generate candidate explanations using $\pi_{\text{sft}}$ and evaluate them using the target LLM's mechanistic activation scores as the objective reward.

We first sample $K_2$ unseen features, denoted as $\mathcal{V}_2 = \{v_j\}_{j=1}^{K_2}$. For each $v_j$, we sample $\pi_{\text{sft}}$ at a high temperature to generate $M$ candidate explanations, $E_j = \{e_j^m\}_{m=1}^M$. To evaluate each candidate $e_j^m$, we prompt a Generator LLM (Appendix \ref{prompt generation}) to synthesize $W$ text samples, $\mathcal{T}_j^m = \{t_1, \dots, t_W\}$, that manifest its semantics. Passing each text $t \in \mathcal{T}_j^m$ through the target LLM, we define the mechanistic score $s(e_j^m)$ as the maximum SAE activation elicited:
\begin{equation}
    s(e_j^m) = \max_{t \in \mathcal{T}_j^m} \left( a_t \right),
\end{equation}
where $a_t$ is the activation value of feature $v_j$ on text $t$.
To ensure high-quality preference pairs, we filter features into a valid set $\mathcal{V}_{\text{valid}}$ based on two criteria: the maximum candidate score must exceed an activation threshold $\tau_{\text{act}}$, and the relative score variance must exceed a margin $\tau_{\text{diff}}$:
\begin{equation}
\left\{
\begin{aligned}
    &\max_{e \in E_j} s(e) \ge \tau_{\text{act}}, \\
    &\frac{\max_{e \in E_j} s(e) - \min_{e \in E_j} s(e)}{\max_{e \in E_j} s(e)} \ge \tau_{\text{diff}}.
\end{aligned}
\right.
\end{equation}

For each valid feature $v_j \in \mathcal{V}_{\text{valid}}$, the positive sample $y_{\text{w}}^{(j)}$ is the highest-scoring candidate: $y_{\text{w}}^{(j)} = \arg\max_{e \in E_j} s(e)$. For the negative sample $y_{\text{l}}^{(j)}$, we first define a subset of candidates $\mathcal{E}_j^{-}$ whose scores are strictly lower than $y_{\text{w}}^{(j)}$ by at least a relative margin $\tau_{\text{margin}}$:
\begin{equation}
    \mathcal{E}_j^{-} = \left\{ e \in E_j \;\middle|\; \frac{s(y_{\text{w}}^{(j)}) - s(e)}{s(y_{\text{w}}^{(j)})} \ge \tau_{\text{margin}} \right\}.
\end{equation}
To construct a challenging negative example, we select the highest-scoring candidate within this subset as $y_{\text{l}}^{(j)}$. If $\mathcal{E}_j^{-}$ is empty, we default to the lowest-scoring candidate:
\begin{equation}
    y_{\text{l}}^{(j)} = 
    \begin{cases} 
        \arg\max_{e \in \mathcal{E}_j^{-}} s(e), & \text{if } \mathcal{E}_j^{-} \neq \emptyset \\ 
        \arg\min_{e \in E_j} s(e). & \text{otherwise} 
    \end{cases}
\end{equation}

Aggregating these pairs yields our preference dataset for DPO training: 
$Y_\textrm{dpo}^{(j)} = (y_{\text{w}}^{(j)}, y_{\text{l}}^{(j)})$, 
$\mathcal{D}_{\text{DPO}} = \left\{ \left(X^j, Y_\textrm{dpo}^{(j)}\right) \right\}_{v_j \in \mathcal{V}_{\text{valid}}}$.

\subsection{Iterative Direct Preference Optimization}
\label{sec:iterative_dpo}

After obtaining the preference dataset, we employ Direct Preference Optimization (DPO) to train the explainer. To continuously improve the model's performance, we conduct two sequential rounds of DPO. DPO implicitly optimizes the reward model by minimizing the following loss function:
\begin{equation}
\begin{aligned}
    &\mathcal{L}_{\text{DPO}}(\pi_\theta; \pi_{\text{ref}}) = -\mathbb{E}_{(x, y_w, y_l) \sim \mathcal{D}} \Bigg[ \log \sigma \Bigg( \\
    &\quad \beta \log \frac{\pi_\theta(y_w|x)}{\pi_{\text{ref}}(y_w|x)} - \beta \log \frac{\pi_\theta(y_l|x)}{\pi_{\text{ref}}(y_l|x)} \Bigg) \Bigg],
\end{aligned}
\end{equation}
where $\pi_\theta$ is the policy model being trained, $\pi_{\text{ref}}$ is the reference model, and $\beta$ controls the KL divergence penalty.

In the first DPO round, both the initial policy and the reference model are initialized from the SFT model (i.e., $\pi_{\text{ref}} = \pi_{\text{sft}}$). We train the model using the initial dataset $\mathcal{D}_{\text{DPO}}^{(1)}$ generated by the $\pi_{\text{sft}}$ explainer. This yields the intermediate model $\pi_{\text{DPO}_1}$, which already demonstrates a substantial improvement in explanation capabilities. 
To further push the performance ceiling, we initiate a second DPO round. We utilize $\pi_{\text{DPO}_1}$ to construct a stronger, on-policy preference dataset $\mathcal{D}_{\text{DPO}}^{(2)}$ following the exact procedure in Section \ref{sec:preference_construction}. By executing the second round of DPO training on $\mathcal{D}_{\text{DPO}}^{(2)}$ with $\pi_{\text{ref}} = \pi_{\text{DPO}_1}$, we obtain our final explainer model.

\section{Experiments}
In this section, we conduct experiments to evaluate the proposed SAEExplainer framework.
\begin{table*}[t]
\centering
\resizebox{\textwidth}{!}{
    \begin{tabular}{l cccccc cccccc}
\toprule

& \multicolumn{6}{c}{\textbf{Layer 1}} & \multicolumn{6}{c}{\textbf{Layer 2}} \\
\cmidrule(lr){2-7} \cmidrule(lr){8-13}

\textbf{Method} 
& \textsc{Gen}~$\uparrow$ & \textsc{Inp}~$\uparrow$ & \textsc{Out}~$\uparrow$ & $A_{\text{high}}$~$\uparrow$ & $A_{\text{low}}$~$\downarrow$ & $\Delta A$~$\uparrow$ 
& \textsc{Gen}~$\uparrow$ & \textsc{Inp}~$\uparrow$ & \textsc{Out}~$\uparrow$ & $A_{\text{high}}$~$\uparrow$ & $A_{\text{low}}$~$\downarrow$ & $\Delta A$~$\uparrow$ \\
\midrule

\rowcolor{gray!15} 
\multicolumn{13}{l}{\textbf{\textit{Target LLM: Gemma-2-9b (SAE: gemmascope-res-16k)}}} \\

Activation Oracles      
& 20.33 & 58.67 & 61.67 & 17.77 & 0.49 & 17.28  
& 26.97 & 64.67 & 62.67 & 47.41 & 1.21 & 46.20 \\

Neuronpedia             
& 21.03 & 62.67 & 62.67 & 19.01 & 0.62 & 18.39  
& 27.77 & 70.33 & 57.33 & 50.13 & 1.42 & 48.71 \\

N-pedia w/ Claude       
& 40.00 & 71.33 & 63.33 & 24.68 & 0.70 & 23.98 
& 49.43 & \textbf{86.33} & 63.00 & 69.01 & \textbf{0.99} & 68.02 \\

N-pedia w/ GPT-5        
& 38.87 & 77.00 & 63.33 & 26.22 & 0.55 & 25.67  
& 49.33 & \textbf{86.33} & 64.00 & 72.46 & 1.42 & 71.04 \\

\rowcolor{cyan!10} 
\textbf{SAEExplainer} 
& \textbf{46.13} & \textbf{77.33} & \textbf{71.00} & \textbf{28.72} & \textbf{0.47} & \textbf{28.25} 
& \textbf{54.50} & 84.33 & \textbf{72.33} & \textbf{75.49} & 1.19 & \textbf{74.30} \\
\midrule

\rowcolor{gray!15} 
\multicolumn{13}{l}{\textbf{\textit{Target LLM: Llama3.1-8b (SAE: llamascope-res-32k)}}} \\

Activation Oracles        
& 31.17 & 41.67 & 70.67 & 3.89 & 0.06 & 3.83  
& 21.77 & 50.66 & 70.00 & 5.00 & \textbf{0.04} & 4.96 \\

Neuronpedia             
& 33.93 & 49.00 & 69.33 & 4.56 & 0.02 & 4.54  
& 27.87 & 56.00 & 69.00 & 5.81 & 0.06 & 5.75 \\

N-pedia w/ Claude       
& 55.67 & 61.33 & 71.66 & 6.52 & 0.03 & 6.49  
& 49.67 & 75.00 & 73.33 & 8.79 & 0.10 & 8.69 \\

N-pedia w/ GPT-5        
& 54.47 & 63.33 & 70.00 & 6.50 & 0.02 & 6.48  
& 46.70 & 77.33 & 75.67 & 8.49 & \textbf{0.04} & 8.45 \\

\rowcolor{cyan!10} 
\textbf{SAEExplainer} 
& \textbf{58.40} & \textbf{68.00} & \textbf{76.00} & \textbf{7.19} & \textbf{0.01} & \textbf{7.18} 
& \textbf{52.33} & \textbf{77.67} & \textbf{77.00} & \textbf{9.08} & 0.09 & \textbf{8.99} \\
\midrule

\rowcolor{gray!15} 
\multicolumn{13}{l}{\textbf{\textit{Target LLM: Gemma-2-27b (SAE: gemmascope-res-131k)}}} \\

Activation Oracles      
& 29.67 & 55.33 & 60.67 & 113.53 & 0.29 & 113.24  
& 10.23 & 27.00 & 35.33 & 102.37 & 1.11 & 101.26 \\

Neuronpedia             
& 37.63 & 67.33 & 64.00 & 132.27 & 0.24 & 132.03  
& 16.17 & 38.33 & 33.00 & 144.01 & 0.61 & 143.40 \\

N-pedia w/ Claude       
& 44.37 & 73.33 & 62.33 & 143.03 & 2.71 & 140.32  
& 18.80 & 42.00 & 39.33 & 158.19 & 5.39 & 152.80 \\

N-pedia w/ GPT-5        
& 35.27 & 72.67 & 62.00 & 132.18 & \textbf{0.13} & 132.05  
& 18.60 & 46.67 & \textbf{40.00} & 177.21 & 4.80 & 172.41 \\

\rowcolor{cyan!10} 
\textbf{SAEExplainer} 
& \textbf{53.63} & \textbf{82.67} & \textbf{67.67} & \textbf{163.69} & \textbf{0.13} & \textbf{163.56} 
& \textbf{26.63} & \textbf{51.00} & 38.33 & \textbf{216.60} & \textbf{0.29} & \textbf{216.31} \\

\bottomrule
\end{tabular}
}
\caption{Quantitative comparison of feature explanation generation methods. The metrics \textsc{Gen}, \textsc{Inp}, and \textsc{Out} denote Generative Accuracy, Input Score, and Output Score, respectively. \textbf{Bold} indicates the best performance, with $\uparrow$ denoting higher is better and $\downarrow$ denoting lower is better.}
\label{tab:layer_comparison_colored}
\end{table*}
\subsection{Experimental Settings}

\paragraph{Models.}
Our experiments are conducted on various target LLMs paired with their respective SAEs: Gemma-2-9B \citep{team2024gemma} using \texttt{gemmascope-res-16k} \citep{lieberum2024gemma}, Gemma-2-27B \citep{team2024gemma} using \texttt{gemmascope-res-131k} \citep{lieberum2024gemma}, and Llama-3.1-8B \citep{grattafiori2024llama} using \texttt{llamascope-res-32k} \citep{he2024llama}. To train our explainer LLMs, we initialize the backbone networks using the instruction-tuned variants of these target models: gemma-2-9b-it, gemma-2-27b-it, and llama-3.1-8b-instruct, respectively. We adopt these instruction-tuned variants because their identical hidden dimensions ensure alignment with the SAE feature vectors.

\paragraph{Baselines.}
To rigorously evaluate our method, we conduct comparisons against the following baselines: (1) Activation Oracles \citep{karvonen2025activation}, which trains models to answer arbitrary questions about LLM activations, and (2) Neuronpedia \citep{neuronpedia}, the current state-of-the-art SAE interpretation platform based on the AutoInterp method \citep{bills2023language}. Furthermore, our comparison with Neuronpedia encompasses not only its default explanations generated by GPT-4o-mini \citep{hurst2024gpt}, but also the high quality explanations from the most advanced models available on the platform, specifically GPT-5 \citep{singh2025openai} and Claude Sonnet 4.5 \citep{anthropic_claude_sonnet_4_5}.

\begin{table*}[t]
    \centering
    \includegraphics[width=\textwidth]{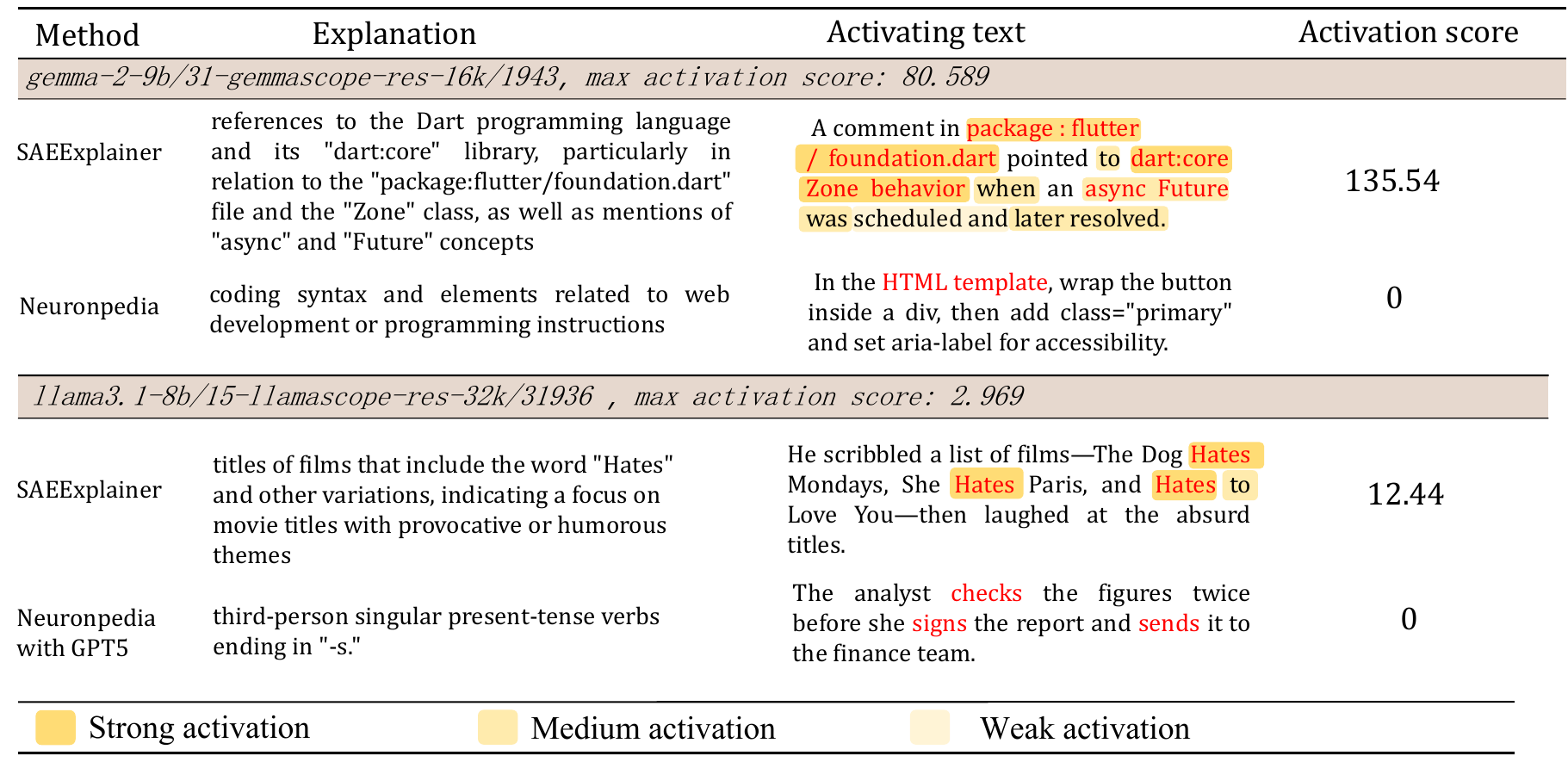}
    \caption{Qualitative case studies with two samples. 
     \textcolor{red}{Red text} highlights matching keywords between explanations and generated texts. Activation levels (Strong/Medium/Weak) represent $>75\%$, $25\%$--$75\%$, and $<25\%$ of the maximum activation, which is retrieved from Neuronpedia's cache.}
    \label{tab:case_study_heatmap}
\end{table*}

\paragraph{Implementation Details.} 
For each target model, we select two specific layers for comparison. In our results in Table \ref{tab:layer_comparison_colored}, we denote these selected layers as Layer 1 and Layer 2. We obtain the initial explanations, as well as those generated by GPT-5 \citep{singh2025openai} and Claude Sonnet 4.5 \citep{anthropic_claude_sonnet_4_5}, by querying the Neuronpedia API \citep{neuronpedia}. Specifically, we employ the default Neuronpedia explanations to construct our training dataset for the SFT phase. We uniformly utilize gpt-5.4-mini \citep{openai2026gpt54mini} for conducting downstream evaluations. During the evaluation phase, we exclusively sample unseen features for experimental testing.
Additional implementation details and hyperparameter settings are provided in the Appendix \ref{app:experiment}.

\paragraph{Evaluation Metrics.}
To evaluate the generated explanations, we employ a comprehensive suite of metrics. First, we use Generative Accuracy \citep{han2026sage} to assess causal validity, verifying if an explanation can guide an LLM to synthesize texts that trigger the target feature. Second, we adopt Input and Output scores \citep{gur2025enhancing} to jointly evaluate the faithfulness and completeness of the explanation, measuring how accurately it captures activating inputs and reflects the feature's impact on model outputs. In addition, to evaluate specificity, we introduce Discriminative Activation Metrics, reporting the average activations on maximally activating ($A_{\text{high}}$) and low-activating ($A_{\text{low}}$) samples, along with their gap ($\Delta A$). A larger $\Delta A$ robustly demonstrates that the explanation captures the precise causal trigger, enforcing clear semantic boundaries rather than vague correlations. 
At last, to evaluate the semantic specificity of feature explanations, we introduce two metrics: Feature Semantic Similarity Rate (FSSR), which compares the geometric relationships among SAE features with the semantic similarity of their corresponding explanations; and Distant Text Similarity, which measures the average explanation similarity of such feature pairs. Lower values indicate better semantic distinction between SAE features.
Detailed implementations are in Appendix \ref{sec:appendixA}.


\subsection{Explanation Results Comparisons}
\label{sec:results_comparison}

Table \ref{tab:layer_comparison_colored} presents the quantitative comparison of our proposed \textbf{SAEExplainer} against various baseline methods across multiple target LLMs and SAE configurations. Overall, SAEExplainer improves upon established baselines across most metrics, demonstrating a strong ability to generate faithful and causally precise explanations. First, in terms of Generative Accuracy, our method consistently surpasses all existing baselines; notably, on Layer 1 of Gemma-2-27b, it outperforms the strongest baseline by a significant margin of 9.26\%. This indicates that the explanations generated by SAEExplainer accurately delineate the input distribution that triggers the features. Furthermore, the fact that texts synthesized based on these explanations reliably elicit high activations in the target model strongly underscores their high accuracy and causal validity. Regarding the Input and Output scores, our method maintains a leading position in the vast majority of cases, exhibiting the strongest comprehensive performance. This demonstrates that the explanations not only faithfully capture the input-side response distribution but also accurately reveal their causal impact on the target model's generative outputs. In addition, SAEExplainer establishes an absolute superiority in the Discriminative Activation Metrics ($\Delta A$). Although it occasionally exhibits minor fluctuations on low-activation samples ($A_{\text{low}}$), these remain well within a reasonable noise tolerance. Meanwhile, it consistently maintains a dominant advantage on high-activation samples ($A_{\text{high}}$). The pronounced $\Delta A$ margin demonstrates that SAEExplainer anchors true causal triggers via activation feedback and suppresses false explanations through contrastive learning, effectively mitigating hallucinations. Crucially, compared to the Neuronpedia data used as our initial SFT training data, our model achieves significant improvements across all evaluated metrics. This further validates the efficacy of our closed-loop feedback mechanism in correcting and refining explanations.

\begin{figure*}[t] 
    \centering
    \begin{subfigure}{0.48\textwidth}
        \centering
        \includegraphics[width=\linewidth]{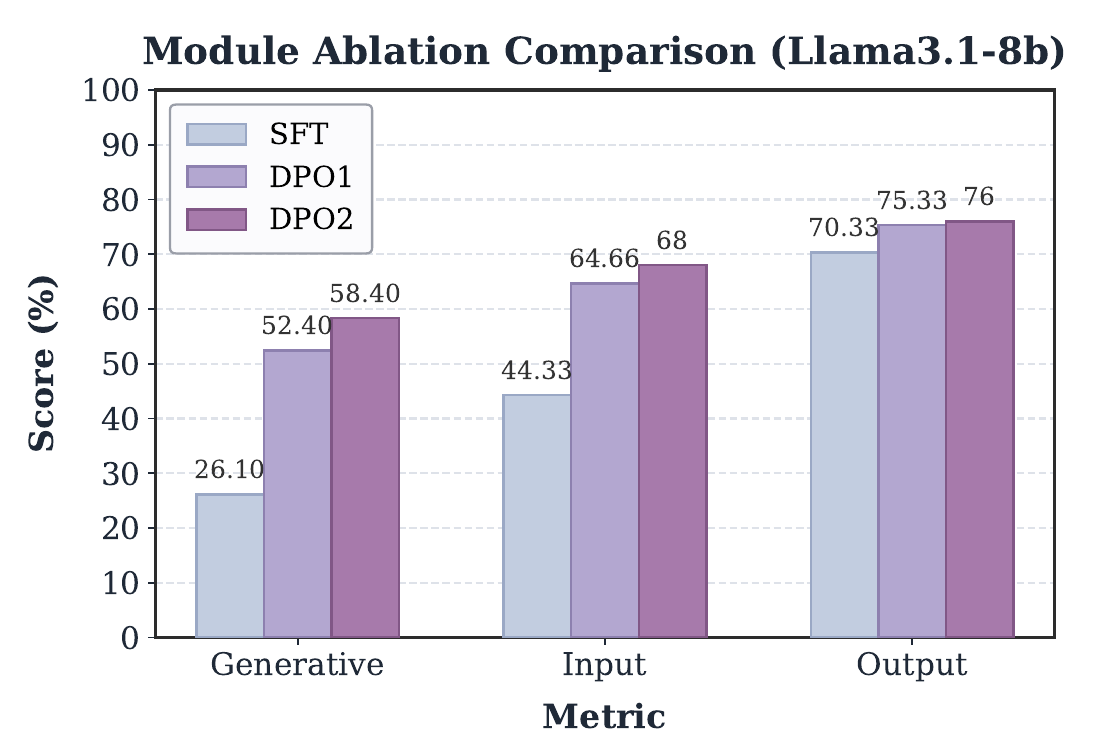}
        \caption{Module Ablation Comparison on Llama3.1-8b}
        \label{fig:sub_ablation}
    \end{subfigure}
    \hfill 
    \begin{subfigure}{0.48\textwidth}
        \centering
        \includegraphics[width=\linewidth]{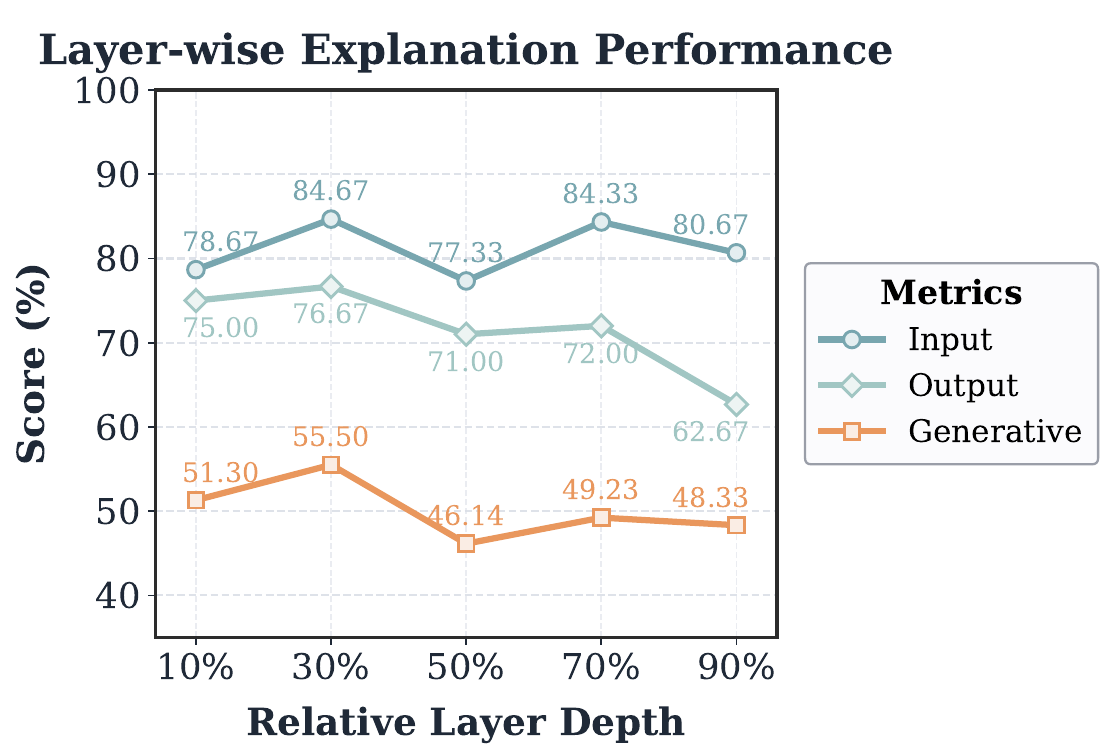}
        \caption{Layer-wise Explanation Performance}
        \label{fig:sub_layer}
    \end{subfigure}
    
    \caption{Module ablation and layer-wise performance analysis of SAEExplainer. 
    (a) Module ablation comparison of SAEExplainer at different stages. 
    (b) Layer-wise performance analysis across different depths of the model.}
    \label{fig:combined_performance}
\end{figure*}


\subsection{Semantic Collision Analysis}
\label{sec:semantic_collision}

To further evaluate the semantic specificity of feature explanations, we conduct a semantic collision analysis by comparing pairwise similarities in the SAE feature space and the explanation space. Intuitively, features with substantially different latent representations should not receive highly similar descriptions. We sample 2,000 held-out features from Layer 15 of Llama-3.1-8B with LlamaScope-32K, and compute cosine similarities between SAE decoder vectors and between explanation embeddings encoded by Qwen3-Embedding-4B \citep{zhang2025qwen3}.
As shown in Table~\ref{tab:semantic_collision}, SAEExplainer substantially reduces semantic collisions compared with both baselines. FSSR decreases from 15.92\% for Neuronpedia to 3.74\%, a 76.5\% relative reduction, while Distant Text Similarity decreases from 0.475 to 0.383. The case studies in Table~\ref{tab:semantic_collision_cases} further illustrate this behavior: for vector-distant features, SAEExplainer provides more specific and distinguishable explanations, whereas Neuronpedia tends to use broader descriptions; for vector-near features, SAEExplainer still preserves reasonable semantic similarity. Together, these results indicate that SAEExplainer better distinguishes distinct SAE features while preserving similarities among related ones, thereby mitigating over-generalization in feature explanations.

\begin{table}[t]
\centering
\small
\begin{tabular}{lcc}
\toprule
\textbf{Method} & \textbf{FSSR $\downarrow$} & \textbf{Distant Text Sim $\downarrow$} \\
\midrule
Neuronpedia & 15.92 & 0.475 \\
N-pedia (Claude) & 6.66 & 0.406 \\
\textbf{SAEExplainer} & \textbf{3.74} & \textbf{0.383} \\
\bottomrule
\end{tabular}
\caption{Results of semantic collision analysis. Lower FSSR and Distant Text Similarity indicate better semantic distinction between different SAE features.}
\label{tab:semantic_collision}
\end{table}

\begin{table*}[t]
\centering
\small
\renewcommand{\arraystretch}{1.18}
\setlength{\tabcolsep}{6pt}
\begin{tabularx}{\textwidth}{@{} p{0.08\textwidth} >{\raggedright\arraybackslash}p{0.55\textwidth} >{\raggedright\arraybackslash}X @{} }
\toprule
\textbf{Feature} & \textbf{SAEExplainer Explanation} & \textbf{Neuronpedia Explanation} \\
\midrule

\rowcolor{gray!10}
\multicolumn{3}{@{}l}{\textbf{Case:} \textit{Vector-Distant Features}} \\
32699
& HTML and CSS elements, especially styling/layout, IDs, and styles for \texttt{body} and \texttt{div} elements.
& Technical terms related to web development and design. \\ \addlinespace[2pt]

14512
& Database seeding and migration code, especially the \texttt{db} object and Laravel \texttt{migrate}/\texttt{seed} functions.
& Technical terms related to databases and software development. \\
\midrule

\rowcolor{gray!10}
\multicolumn{3}{@{}l}{\textbf{Case:} \textit{Vector-Near Features}} \\
9706
& Instances of the verb ``has'' in various contexts, indicating a focus on statements or assertions about possession or existence.
& Phrases indicating possession or ownership. \\ \addlinespace[2pt]

18653
& Instances of the verb ``have'' in various forms, particularly focusing on ``have'' and ``had'' in contexts suggesting possession or existence.
& Instances of the word ``have'' in various forms and contexts. \\

\bottomrule
\end{tabularx}
\caption{Representative case studies comparing SAEExplainer and Neuronpedia explanations for vector-distant and vector-near SAE feature pairs.}
\label{tab:semantic_collision_cases}
\end{table*}
\subsection{Qualitative Evaluation}
In this section, we qualitatively demonstrate that for specific feature explanations, our Explainer effectively alleviates the overly broad semantics and hallucinations exhibited by the baselines. Table \ref{tab:case_study_heatmap} presents two case studies that highlight our method's exceptional precision and faithfulness.

The first example (Feature 1943) reveals the over-generalization problem in the existing baseline. While Neuronpedia broadly interprets the feature as web development instructions, texts generated from this explanation yield zero activation, indicating a merely superficial topical definition. In contrast, our SAEExplainer pinpoints the precise sub-domain (Dart programming language) and explicitly identifies core objects like \texttt{package:flutter} and \texttt{Zone}. As a result, texts synthesized from these explanations achieve remarkably high activation scores, rigorously validating SAEExplainer's causal validity.
The second case (Feature 31936) illustrates our method's robust ability to resist explanation hallucination. The baseline incorrectly abstracts the feature into a plausible but spurious concept: third-person singular verbs. Consequently, its corresponding samples fail to activate the feature. SAEExplainer avoids this false abstraction and correctly identifies the exact trigger as the word ``Hates'' within film titles. Remarkably, the maximum activation scores in our two samples even surpass the original maximum activation score retrieved from Neuronpedia, definitively confirming that our approach transcends the limitations of external proxies to achieve self-improvement.



\subsection{Layer-wise Performance Analysis}
\label{sec:layer_wise_evaluation}
Our results demonstrate that SAEExplainer maintains faithfulness and causal triggering capabilities across sampled layers representing early, middle, and late depths, successfully deciphering both concrete early-layer concepts and highly abstract deep-layer representations. To evaluate this, we analyze Gemma-2-9B (with the \textit{gemmascope-res-16k} SAE) at relative layer depths of 10\%, 30\%, 50\%, 70\%, and 90\% (i.e., layers 4, 12, 20, 29, and 37).

The results in Figure \ref{fig:sub_layer} demonstrate a non-linear performance trajectory across different depths. First, all three explanation evaluation metrics reach their highest values at the 30\% relative depth. This simultaneous peak indicates that the explainer model achieves its optimal performance here, successfully capturing both the trigger conditions and the downstream impacts of the features.
Subsequently, performance across all metrics experiences a distinct dip at the 50\% layer. This synchronized decline highlights a localized challenge for the explainer, suggesting that intermediate layer features are inherently more difficult to articulate.
Furthermore, from the 70\% to 90\% layers, a notable metric divergence occurs: the Input score rebounds and remains stable above 80\%, whereas the Output score exhibits a continuous decline, reaching its minimum at the 90\% depth. This divergence reveals an important property of deep-layer explanations: while SAEExplainer can still effectively identify the upstream trigger conditions of a feature, it becomes increasingly difficult to accurately map how the feature influences the model's final predictions.


\subsection{Ablation Study}
\label{sec:ablation_study}

Our ablation study demonstrates that using activation scores as rewards overcomes the explanatory limits of pure label imitation. Furthermore, iterative DPO surpasses single-pass optimization, significantly reducing hallucinations and reinforcing the fidelity of causal triggers. To systematically evaluate the contributions of these core modules in SAEExplainer, we analyzed Llama3.1-8b (using the llamascope-res-32k SAE, targeting layer 15 features) across three training stages: SFT, the first round of DPO (DPO1), and the second round (DPO2), utilizing Generative Accuracy, Input Score, and Output Score as key metrics.
As shown in Figure \ref{fig:sub_ablation}, the SFT-only model ranks lowest across all metrics. However, introducing DPO1 yields a significant leap: Generative Accuracy surges by ~26\%, while Input and Output Scores achieve absolute gains of 20.33\% and 5\%, respectively. This validates the effectiveness of our objective reward mechanism. By utilizing activation scores for preference alignment, DPO forces the model to strictly delineate triggering boundaries between positive and negative samples. Following DPO2, all metrics sustain further improvements. This confirms our self-evolution strategy: as explanatory capability enhances, the model synthesizes higher-quality preference datasets for the next DPO round, creating a positive feedback loop between model capability and data quality. Furthermore, Table \ref{tab:explanation_evolution_combined} demonstrates how explanations evolve across training stages using examples from two different models. Initial SFT explanations remain overly broad and deviate from actual triggers. While DPO1 begins to correct and explore these concepts, DPO2 provides the crucial refinement needed to capture the exact causal triggers of feature activation, ultimately achieving a perfect generative accuracy of 1.0.

\begin{table*}[t!]
\centering
\small
\definecolor{hlpink}{RGB}{210, 50, 120}

\begin{tabularx}{\textwidth}{@{} l >{\raggedright\arraybackslash}X c @{}}
\toprule
\textbf{Stage} & \textbf{Explanation} & \textbf{Generative} \\ \midrule

\rowcolor{gray!10} \multicolumn{3}{@{}l}{\textbf{Feature:} \textit{llama3.1-8b/15-llamascope-res-32k/9386}} \\
SFT            & references to naval ships and their specifications & 0.0 \\ \addlinespace
DPO1           & references to shipbuilding and naval history, particularly related to ship classes and their fates & 0.8 \\ \addlinespace
DPO2           & references to shipbuilding and \textcolor{hlpink}{commissioning}, particularly for ships of the \textcolor{hlpink}{Royal Navy}, including \textcolor{hlpink}{cancellation} and fate information & \textbf{1.0} \\ \midrule

\rowcolor{gray!10} \multicolumn{3}{@{}l}{\textbf{Feature:} \textit{gemma-2-9b/20-gemmascope-res-16k/16352}} \\
SFT            & HTML tags and their attributes & 0.0 \\ \addlinespace
DPO1           & HTML tags, specifically the opening tag for a table footnote & 0.0 \\ \addlinespace
DPO2           & HTML tag representations of \textcolor{hlpink}{superscript}, specifically the \textcolor{hlpink}{opening and closing} tags \texttt{\textcolor{hlpink}{<sup>}} and \texttt{\textcolor{hlpink}{</sup>}} & \textbf{1.0} \\ \bottomrule
\end{tabularx}
\caption{Comparison of feature explanations generated across different training stages for distinct SAE features. Key improvements in specificity by the final phase (DPO2) are highlighted in \textcolor{hlpink}{pink}.}
\label{tab:explanation_evolution_combined}
\end{table*}


\section{Related Work}
\paragraph{Mechanistic Interpretability and SAEs.}
Mechanistic interpretability seeks to reverse-engineer internal network computations into human-understandable concepts \citep{elhage2021mathematical}. However, dense LLM representations cause severe neuron polysemanticity, hindering this direct interpretation \citep{elhage2022superposition}. To address this, prior work introduced SAEs, which project dense activations into higher-dimensional sparse latent spaces and thereby recover more nearly monosemantic features \citep{bricken2023monosemanticity,huben2024sparse}. Subsequent work studied the scaling behavior of SAEs \citep{gao2025scaling} and proposed improved variants such as Gated SAEs to further improve reconstruction fidelity \citep{rajamanoharan2024improving}. More recently, Gemma Scope has made millions of high-quality SAE features available to the community at scale \citep{lieberum2024gemma}. However, as feature inventories grow, the primary bottleneck has shifted from feature discovery to automatically generating high-fidelity natural language explanations.

\paragraph{Automated Interpretability.}
To address the unscalability of manual annotation, automated interpretability has emerged as a vital direction in mechanistic interpretability research. Pioneering early works, such as OpenAI's AutoInterp, utilized LLMs to read top-activating text snippets of neurons to generate explanations \citep{bills2023language} and subsequently scaled to massive SAE features \citep{paulo2024automatically}.
Concurrently, LatentQA and Activation Oracles attempt to answer natural language questions directly from internal activations \citep{pan2026latentqa,karvonen2025activation}. At the platform level, Neuronpedia has provided essential infrastructure for the retrieval and sharing of neuron and SAE features \citep{neuronpedia}.
While these methodologies have advanced the comprehension of SAE features, they are limited to open-loop generation, suffering from widespread explanation hallucinations and over-generalization.

\section{Conclusions}
In this work, we note that current approaches lack mechanisms to leverage objective feedback for explanation refinement, rendering hallucinations and over-generalization widespread. To address these limitations, we introduced SAEExplainer, a novel framework that  utilizes
activation scores as an objective reward signal to train the model for self-correction and iterative improvement. 
Comprehensive evaluations demonstrate that SAEExplainer generates highly faithful explanations with superior causal triggering properties compared to existing baselines. Furthermore, SAEExplainer demonstrates effective performance across the evaluated language models and sampled network depths, substantiating its broad applicability across different architectures and representation stages.

\section*{Limitations}
While our proposed SAEExplainer framework advances automated SAE feature explanation, we acknowledge several limitations. First, compared to conventional single-pass generation methods, our approach necessitates additional computational resources and time overhead. It might introduce a scalability challenge when deploying the framework on extremely large-scale models under constrained hardware budgets. Second, to ensure strict dimensional alignment during the feature vector injection phase, we currently restrict our explainer to the instruction-tuned variant of the target model's architecture. The effectiveness and training stability of our framework in cross-architecture scenarios, where the target model and the explainer possess mismatched hidden dimensions, remain unexplored and are left for future work.

\bibliography{custom}

\clearpage
\appendix

\section{Metrics Details}
\label{sec:appendixA}

In this section, we provide the detailed implementation configurations for the evaluation metrics.
Let $f_j$ denote the target SAE feature ($v_j$ is the corresponding feature vector), and $e_j$ denote the generated explanation for this feature.

\subsection{Generative Accuracy}
Generative Accuracy measures the causal validity of an explanation by testing if it can reliably guide the generation of activating texts. Specifically, we prompt a Generator LLM to synthesize a set of $N$ (in the specific setting, $N=10$) novel sentences designed to strongly activate the target feature $f_j$, based solely on the provided feature explanation $e_j$. (Prompt sees Appendix \ref{Prompt generative}) Notably, to mitigate the risk of evaluation leakage, where the Explainer might overfit to the preference construction pipeline, we strictly isolated the prompts used for this evaluation from those used during preference text generation (Appendix \ref{prompt generation}).
To determine a successful activation, we define a dynamic success threshold $T_{\text{act}}$. This threshold is set to 50\% of the cached maximum activation score for the given feature, which is retrieved via Neuronpedia.
For each generated sentence $s_k$, we extract its maximum token-level activation. The Generative Accuracy is then computed as the fraction of sentences whose maximum activation successfully exceeds the threshold $T_{\text{act}}$:
\begin{equation}
    \text{Generative Accuracy} = \frac{1}{N} \sum_{k=1}^{N} \mathbb{I} \left[ \max_{t \in s_k} f_j(t) > T_{\text{act}} \right]
\end{equation}
where $\mathbb{I}[\cdot]$ is the indicator function, and $f_j(t)$ represents the activation score of feature $f_j$ on token $t$ within the sentence $s_k$.

\subsection{Input and Output Scores}
Adopted from recent automated interpretability benchmarks, the Input and Output scores evaluate the bidirectional alignment between the text space and the latent activation space.

\paragraph{Input Score Evaluation.}
Following the methodology established by \citet{huang2023rigorously}, the Input evaluation measures how accurately the generated explanation $e_j$ captures the specific inputs that trigger the target feature $f_j$.
Given the feature explanation $e_j$, we prompt an evaluator LLM to generate two distinct sets of $k$ text examples ($k=5$): an \textit{activating} set and a \textit{neutral} set (Appendix \ref{prompt input}). The activating examples are logically expected to trigger the feature according to $e_j$, whereas the neutral examples are not.
We then pass these generated examples through the target model to obtain the activation of feature $f_j$. Following prior work which demonstrates that strong, localized activations are the most meaningful indicators of feature presence \citep{bills2023language, choi2024automatic, voita2024neurons}, we calculate the activation score for an entire sequence as the maximum activation over all its token positions.
Let $\bar{m}_{\text{activating}}$ and $\bar{m}_{\text{neutral}}$ denote the mean activations obtained for the activating and neutral example sets, respectively:
\begin{equation}
    \bar{m}_{\text{activating}} = \frac{1}{k} \sum_{x \in \mathcal{S}_{\text{act}}} \max_{t \in x} f_j(t),
\end{equation}
\begin{equation}
    \bar{m}_{\text{neutral}} = \frac{1}{k} \sum_{x \in \mathcal{S}_{\text{neu}}} \max_{t \in x} f_j(t),
\end{equation}
where $\mathcal{S}_{\text{act}}$ and $\mathcal{S}_{\text{neu}}$ represent the sets of activating and neutral examples, and $t$ represents the individual tokens within the text sequence $x$.
The explanation $e_j$ is considered faithful if the mean activation for the activating examples strictly exceeds that of the neutral examples:
\begin{equation}
    \bar{m}_{\text{activating}} > \bar{m}_{\text{neutral}}.
\end{equation}

Essentially, this metric automates the assessment of how well the textual description delineates the precise input distribution that activates the feature.

\paragraph{Output Score Evaluation.}
The Output evaluation assesses how faithfully the description $e_j$ captures the target feature $f_j$'s influence on the model's generative outputs. We evaluate $e_j$ by comparing the texts generated when steering the target feature $f_j$ against those generated when steering random control features.
Concretely, we feed the target LLM with open-ended prompts (e.g., ``I think'') \citep{chalnev2024improving} and allow it to generate $n$ tokens under three distinct intervention conditions:
\begin{enumerate}
    \item Amplifying the target feature $f_j$.
    \item Amplifying a random control feature $f'$.
    \item Amplifying another random control feature $f''$.
\end{enumerate}
Feature amplification is achieved by clamping the feature's activation to a high constant value $m$ during the forward pass. To ensure the intervention is effective yet non-destructive, we follow the practice of running each input with varying amplification levels. More specifically, for a small set of target KL magnitudes, we search for corresponding steering amplitudes that produce the desired intervention strength \citep{bhalla2024towards}. The steering strength is calibrated using the KL divergence between the steered and non-steered output distributions, averaged over all open ended prompts.
This procedure yields three sets of generated texts: $\mathcal{T}_{f_j}$ (steered by the target feature), $\mathcal{T}_{f'}$, and $\mathcal{T}_{f''}$ (steered by random features). 
Next, we concatenate the target explanation $e_j$ with the three sets of generated texts ($\mathcal{T}_{f_j}$, $\mathcal{T}_{f'}$, $\mathcal{T}_{f''}$) in a randomized order and provide them to an evaluator (judge) LLM. The judge LLM is tasked with identifying which of the three text sets best matches the provided description $e_j$ (Appendix \ref{prompt output}). 

The explanation $e_j$ is considered faithful if the judge LLM successfully selects the text set $\mathcal{T}_{f_j}$ generated by amplifying the target feature $f_j$. This evaluation acts as a multiple-choice reading comprehension task for the judge LLM, explicitly measuring how accurately the description captures the feature's causal impact on the model's output generation.

\subsection{Discriminative Activation Metrics}
To evaluate the boundary clarity of the SAE feature explanation and quantitatively assess the specificity of a given description, we introduce the Discriminative Activation Metrics. A higher difference indicates that the explanation can more accurately reflect the specific, intrinsic meaning of the explained feature, rather than providing a vague and generalized topical summary.
We instruct the Generator LLM to synthesize two distinct sets of text samples conditioned on the feature explanation $e_j$ (Appendix \ref{prompt input}):
\begin{enumerate}
    \item \textbf{High-Activation Samples ($\mathcal{S}_{\text{high}}$):} Texts synthesized based on the explanation $e_j$ that are expected to elicit high activation scores within the target model.
    \item \textbf{Low-Activation Samples ($\mathcal{S}_{\text{low}}$):} Texts synthesized according to $e_j$ that are explicitly expected to yield low activation scores in the target model.
\end{enumerate}
We define the High Activation ($A_{\text{high}}$) and Low Activation ($A_{\text{low}}$) as the average maximum token-level activation scores across these respective sets:
\begin{equation}
    A_{\text{high}} = \frac{1}{|\mathcal{S}_{\text{high}}|} \sum_{x \in \mathcal{S}_{\text{high}}} \max_{t \in x} f_j(t)
\end{equation}
\begin{equation}
    A_{\text{low}} = \frac{1}{|\mathcal{S}_{\text{low}}|} \sum_{x \in \mathcal{S}_{\text{low}}} \max_{t \in x} f_j(t)
\end{equation}
where $t$ represents the individual tokens within the text sequence $x$, and $f_j(t)$ denotes the activation magnitude of the target feature.
The discriminative activation gap ($\Delta A$), representing the discriminative margin, is formulated as:
\begin{equation}
    \Delta A = A_{\text{high}} - A_{\text{low}}
\end{equation}

A higher $\Delta A$ robustly demonstrates that the explanation possesses a precise causal trigger and sharp semantic boundaries, effectively isolating the true semantic concept of the feature from spurious correlations.

\subsection{Feature Semantic Similarity Rate}
\label{app:fssr}

Feature Semantic Similarity Rate (FSSR) measures whether an explainer assigns overly similar explanations to SAE features that are substantially different in the feature space. Specifically, for each pair of SAE features, we compute the cosine similarity between their decoder vectors and the cosine similarity between their corresponding explanation embeddings. We define the bottom 10\% of feature pairs ranked by SAE-vector similarity as the set of vector-distant pairs, denoted by $\mathcal{D}$. FSSR is then computed as the fraction of vector-distant pairs whose explanation similarity falls within the top 10\% among all feature pairs:
\begin{equation}
    \mathrm{FSSR}
    =
    \frac{
        \left|
        \left\{
        (i,j) \in \mathcal{D} :
        s^{\mathrm{text}}_{ij} \geq q^{\mathrm{text}}_{90}
        \right\}
        \right|
    }{
        |\mathcal{D}|
    }
    \times 100\%.
\end{equation}
Here, $s^{\mathrm{text}}_{ij}$ denotes the cosine similarity between the explanation embeddings of features $i$ and $j$, and $q^{\mathrm{text}}_{90}$ denotes the 90th percentile of explanation similarity across all feature pairs. A lower FSSR indicates fewer cases in which features that are distant in the SAE representation space receive highly similar explanations, corresponding to fewer semantic collisions.

\subsection{Distant Text Similarity}
\label{app:distant_text_sim}

Distant Text Similarity measures the average semantic similarity between explanations of vector-distant SAE features. Using the same set of vector-distant pairs $\mathcal{D}$, we compute:
\begin{equation}
    \mathrm{DistantTextSim}
    =
    \frac{1}{|\mathcal{D}|}
    \sum_{(i,j) \in \mathcal{D}}
    s^{\mathrm{text}}_{ij}.
\end{equation}
A lower Distant Text Similarity indicates that the explainer produces more distinguishable natural-language descriptions for features that are distant in the SAE representation space, thereby reducing over-generalization in feature explanations.

\begin{table*}[t!]
    \centering
    \footnotesize 
    \setlength{\tabcolsep}{7.5pt} 
    \renewcommand{\arraystretch}{1.1} 
    \begin{tabular}{@{} l cccc cccc ccc @{}}
        \toprule
        & \multicolumn{4}{c}{\textbf{Layer 1}} & \multicolumn{4}{c}{\textbf{Layer 2}} & \multicolumn{3}{c}{\textbf{Shared Settings}} \\
        \cmidrule(lr){2-5} \cmidrule(lr){6-9} \cmidrule(lr){10-12}

        & Idx & $K_1$ & $K_2$ & $\tau_{\text{act}}$ 
        & Idx & $K_1$ & $K_2$ & $\tau_{\text{act}}$ 
        & $N$ & Batch & Grad Accum \\
        \midrule
        
        \rowcolor{cyan!10} 
        \multicolumn{12}{@{}l}{\textbf{\textit{Target LLM: Gemma-2-9b (SAE: gemmascope-res-16k) Explainer: Gemma-2-9b-it}}} \\
        ~ & 20 & 10000 & 5000 & 10 & 31 & 10000 & 5000 & 15 & 16,384 & 4 & 4 \\
        \midrule
        
        \rowcolor{cyan!10} 
        \multicolumn{12}{@{}l}{\textbf{\textit{Target LLM: Llama3.1-8b (SAE: llamascope-res-32k) Explainer: Llama3.1-8b-Instruct}}} \\
        ~ & 15 & 20000 & 10000 & 1 & 25 & 20000 & 10000 & 5 & 32,768 & 4 & 4 \\
        \midrule
        
        \rowcolor{cyan!10} 
        \multicolumn{12}{@{}l}{\textbf{\textit{Target LLM: Gemma-2-27b (SAE: gemmascope-res-131k) Explainer: Gemma-2-27b-it}}} \\
        ~ & 10 & 40000 & 10000 & 50 & 22 & 20000 & 10000 & 100 & 131,072 & 2 & 8 \\
        
        \bottomrule
    \end{tabular}
    \caption{Hyperparameter Configurations for Different Target LLMs}
    \label{tab:hyperparameters}
\end{table*}

\section{Details on Experiments}\label{app:experiment}
All experiments were conducted on a single NVIDIA A100 GPU. Specifically, training the Gemma-2-9B-It and Llama-3.1-8B-Instruct models required 40 GB of VRAM, whereas the larger Gemma-2-27B-It model necessitated an 80 GB VRAM configuration. For the 8B and 9B models, both the supervised fine-tuning (SFT) and a single round of DPO training take approximately one hour to complete. In contrast, constructing the preference dataset is more time-intensive, requiring roughly three to four hours per round. For the 27B model, all aforementioned computational time requirements are approximately doubled. Regarding the hyperparameters introduced in Section \ref{sec:preference_construction}, we uniformly set $M=6$, $W=4$, $\tau_{\text{diff}}=0.1$, and $\tau_{\text{margin}}=0.15$ across all experiments. Given the variance in activation scores across different models and layers, we adaptively established the threshold $\tau_{\text{act}}$ based on their empirical activation distributions, as detailed in Table \ref{tab:hyperparameters}. For different target models, the specific layer selection is also shown in Table \ref{tab:hyperparameters}. For dataset partitioning, the data allocation for the SFT and DPO stages was scaled according to the total volume of each respective dataset. Consistently, the second round of DPO reused the exact subset of features from the first DPO round to construct the new preference dataset, with the sole exception of the 10-th layer of Gemma-2-27B, where we sampled 10,000 new features for the second round. For the evaluation phase, we uniformly selected a hold-out set of 300 unseen features per layer for all assessments.

During the SFT stage, we uniformly applied a learning rate of 5e-5. The LoRA configuration was standardized with $r=32$, $\alpha=64$, a dropout rate of 0.05, and a warmup ratio of 0.03. In the preference data generation phase, we set the sampling temperature to 0.9 to generate multiple candidate explanations. For the first round of DPO training, we set $\beta=0.1$ and the learning rate to 5e-6; in the second round of DPO training, we adjusted $\beta$ to 0.15 and the learning rate to 3e-6.

\begin{figure*}[t]
    \centering
    
    \begin{subfigure}{0.48\textwidth}
        \centering
        \includegraphics[width=\linewidth]{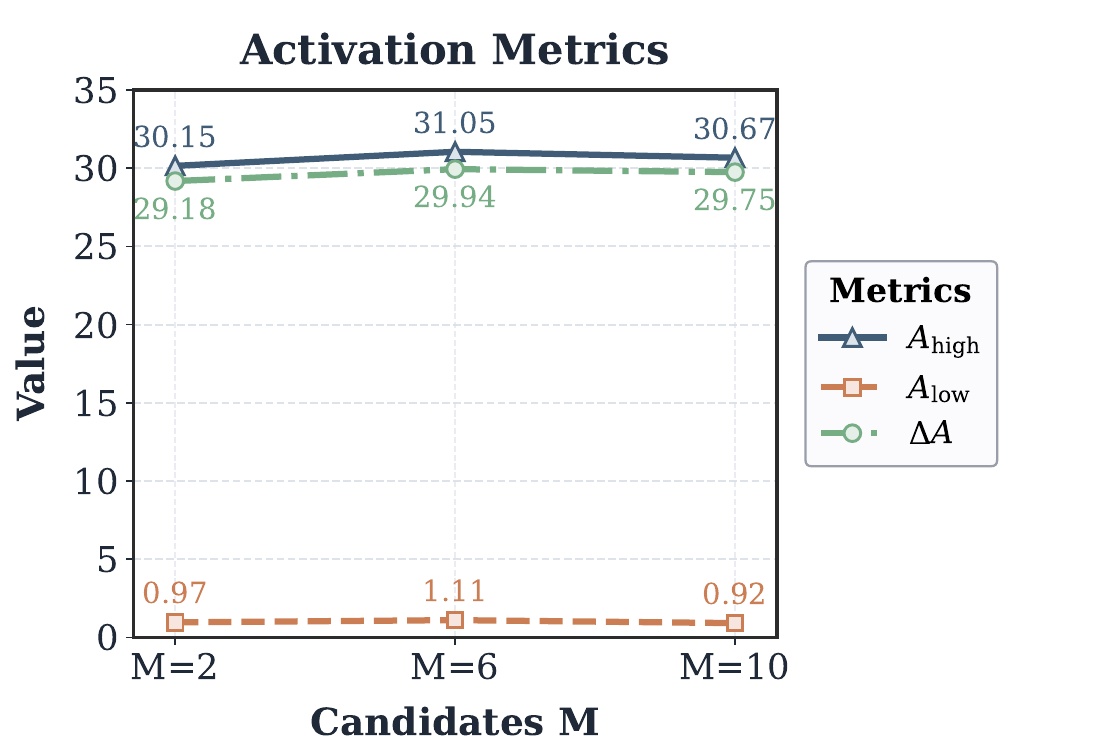}
        \caption{Activation Metrics}
        \label{fig:candidate_m1}
    \end{subfigure}
    \hfill 
    \begin{subfigure}{0.48\textwidth}
        \centering
        \includegraphics[width=\linewidth]{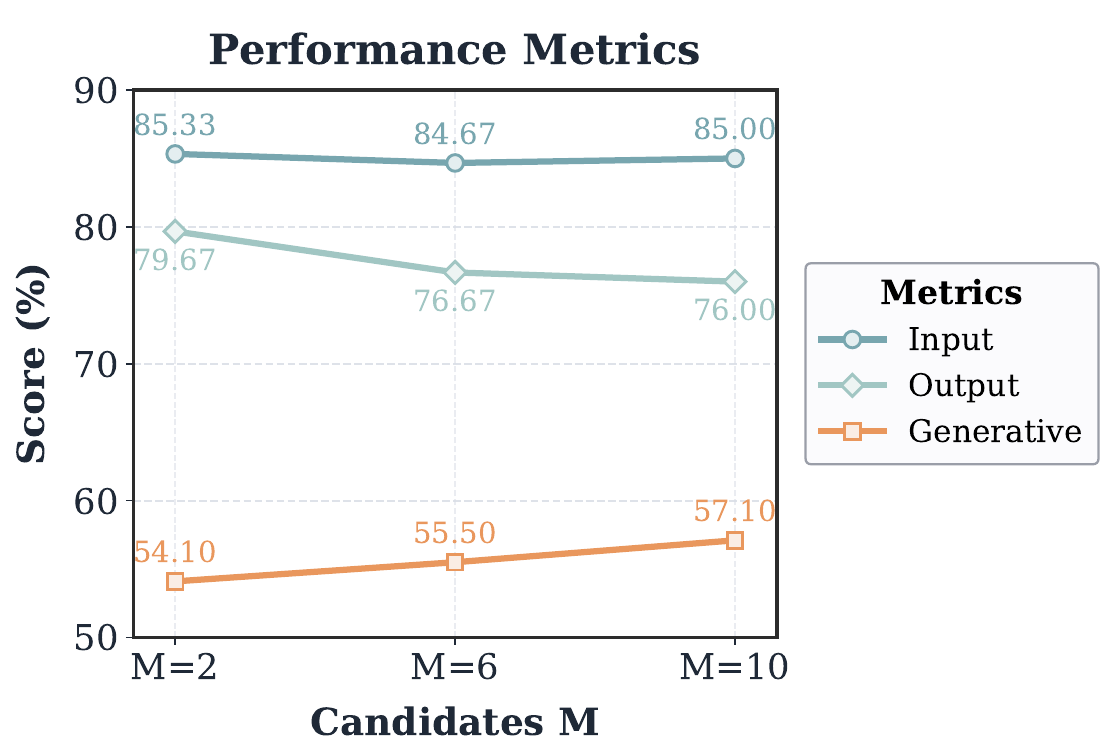}
        \caption{Performance Metrics} 
        \label{fig:candidate_m2}
    \end{subfigure}
    
    \caption{Ablation study on the candidate pool size $M$. The left figure shows the performance of activation score; the right figure shows performance on generative accuracy, input score and output score.}
    \label{fig:ablation_m}

    \begin{subfigure}[b]{0.48\textwidth}
        \centering
        \includegraphics[width=\linewidth]{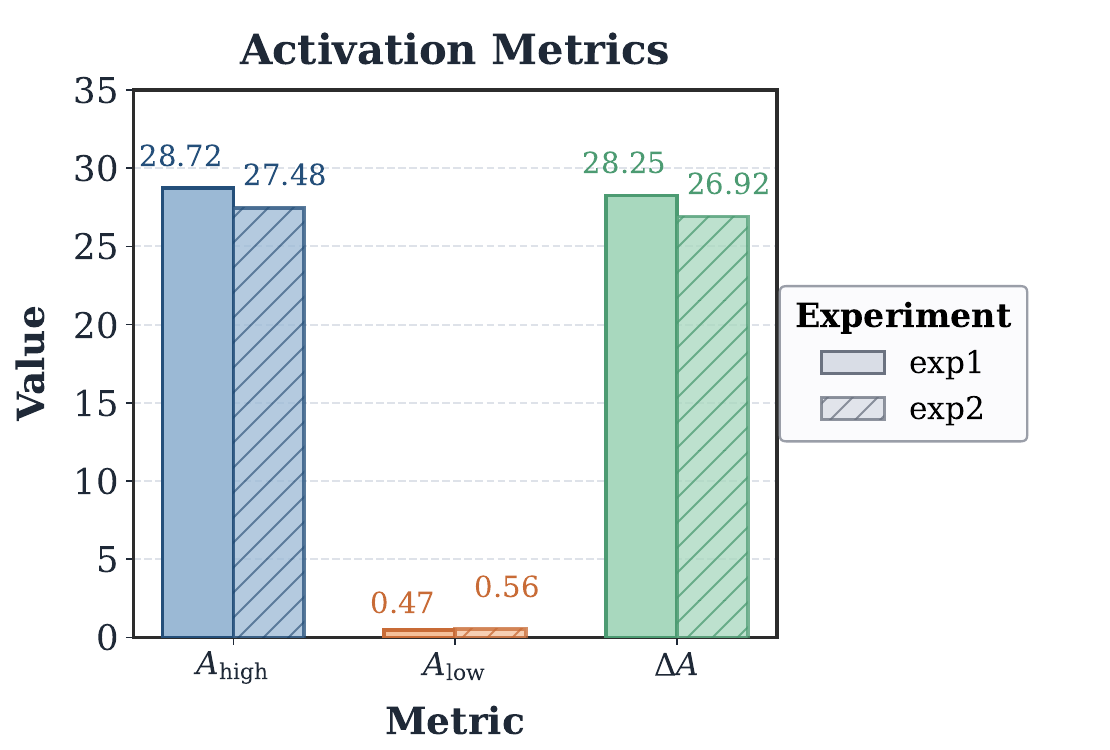}
        \caption{Activation Metrics}
        \label{fig:data_split_activation}
    \end{subfigure}
    \hfill 
    \begin{subfigure}[b]{0.48\textwidth}
        \centering
        \includegraphics[width=\linewidth]{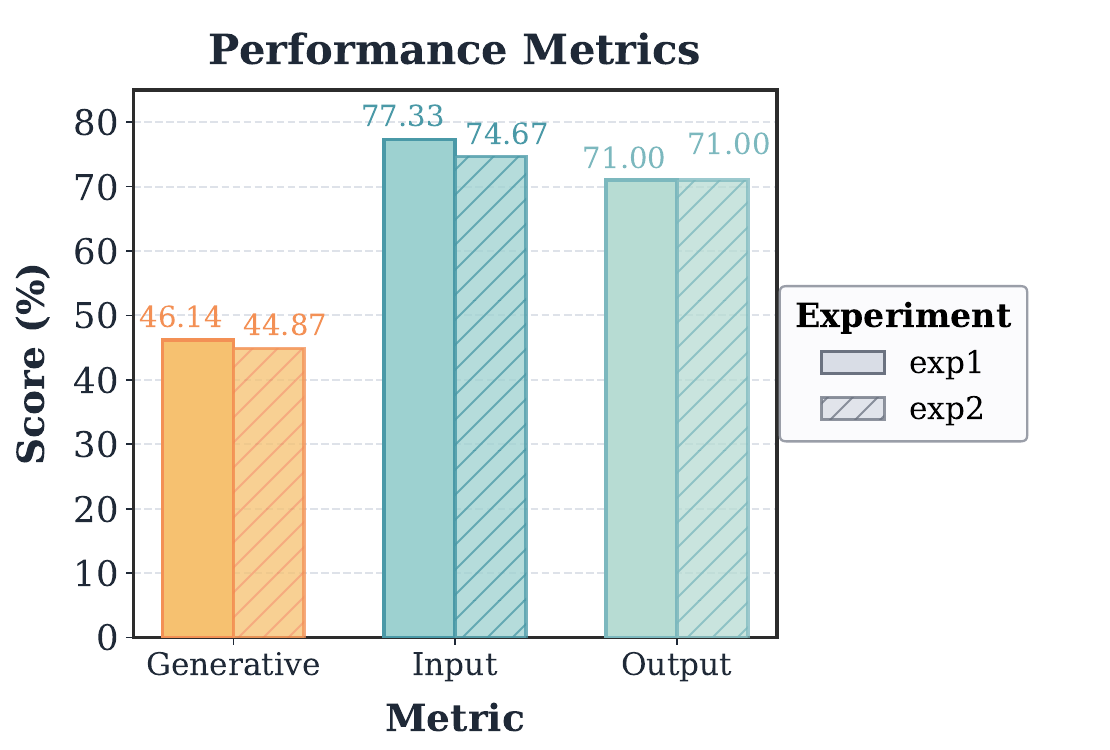}
        \caption{Performance Metrics}
        \label{fig:data_split_performance}
    \end{subfigure}
    
    \caption{Comparison of different data splits. The left figure shows the performance of activation scores; the right figure shows the performance on generative accuracy, input score, and output score.}
    \label{fig:data_split_comparison}
    
\end{figure*}

\section{Ablation Study}
\subsection{Ablation on Candidate Pool Size $M$}
To investigate the parameter sensitivity of the SAEExplainer framework during the preference dataset construction phase, we conducted an ablation study on $M$, the number of candidate explanations generated for each feature. Using gemma-2-9b as the target model and extracting data from layer 12, we evaluated three settings: $M \in \{2, 6, 10\}$ (while keeping all other hyperparameters constant). We comprehensively assessed performance using the Input Score, Output Score, Generative Accuracy, and discriminative activation metrics ($A_{\text{high}}$, $A_{\text{low}}$, and $\Delta A$). Overall, the model's metrics remain relatively stable across different $M$ settings, indicating the framework's fundamental robustness to the candidate quantity parameter. As observed from Figure \ref{fig:ablation_m}, Generative Accuracy exhibits a monotonically increasing trend as $M$ scales from 2 to 10. This suggests that a larger sampling space increases the probability of identifying high-quality positive samples capable of triggering extremely high activation scores. Conversely, the Output Score shows a continuous decline. This may occur because when the candidate pool is strictly limited, the model is forced to carefully distill genuine semantics from limited options, which preserves the feature's global semantics in a way, thereby achieving the best performance in the Output evaluation. However, as $M$ expands to 6, the model becomes more inclined to select explanations with higher activation scores as positive samples. This tendency inherently narrows the semantic scope and compromises the integrity of the global semantics. Nevertheless, the Input Score remains relatively stable, indicating that explanations containing specific trigger words still possess strong sufficiency. Regarding the activation score evaluation, when $M=6$, both the high-activation mean ($A_{\text{high}}$) and the discriminative margin between positive and negative samples ($\Delta A$) reach their peaks ($\Delta A = 29.94$). When $M$ is further increased to 10, although Generative Accuracy continues to improve, $\Delta A$ experiences a decline. This indicates that an excessively large candidate pool may narrow the discriminative gap between positive and negative samples, introducing semantic noise. The physical activation boundary between positive and negative samples becomes blurred, weakening the model's discriminative capability during contrastive learning. Synthesizing the dynamic changes across multidimensional metrics, we observe a fundamental trade-off: strictly optimizing for maximum activation feedback inherently imposes a penalty on broad semantic fidelity. Overall, $M=6$ yields the best comprehensive performance; therefore, we adopt $M=6$ for all subsequent experiments.

\begin{table}[t]
    \centering
    \small 
    \begin{tabular}{lcc}
        \toprule
        & \multicolumn{2}{c}{\textbf{Data Size (\# Features)}} \\
        \cmidrule(lr){2-3} 
        \textbf{Phase} & \textbf{exp1} & \textbf{exp2} \\
        \midrule
        SFT  & 10,000 & 5,000 \\
        DPO1 & 5,000  & 5,000 \\
        DPO2 & 5,000 (Reused) & 5,000 (New) \\
        \midrule
        Total Used / $N$ & 15,000 / 16,384 & 15,000 / 16,384 \\
        \bottomrule
    \end{tabular}
    \caption{Data split configurations for the SFT and iterative DPO phases. Both experiments utilize a total of 15,000 unique features from the available pool ($N=16,384$), differing strictly in their stage-wise allocation and data reuse strategies.}
    \label{tab:data_split_settings}
\end{table}
\subsection{Ablation on dataset split}
In this section, utilizing layer 20 of Gemma-2-9b (SAE: gemmascope-res-16k), we investigate the impact of data allocation and reuse strategies across different training stages on the final explanation performance, given a fixed total budget of 15,000 unique features. As shown in Table \ref{tab:data_split_settings}, we compare exp1, a strategy focused on building a robust SFT foundation while reusing preference data in the subsequent two iterative DPO rounds, against exp2, a balanced strategy that sacrifices SFT scale to introduce entirely new feature data during the DPO phase. The quantitative results in Figure \ref{fig:data_split_comparison} demonstrate that the SFT-heavy and data-reusing strategy (exp1) comprehensively outperforms the new-data-introducing strategy (exp2): on activation metrics, exp1 achieves a higher positive activation ($A_{\mathrm{high}}=28.72$) and lower low-frequency noise ($A_{\mathrm{low}}=0.47$), exhibiting more precise explanatory capabilities and stronger suppression of explanation hallucinations with a significant discriminative margin ($\Delta A$) of 28.25; simultaneously, on comprehensive performance, its generative accuracy (46.14\%) and input score (77.33\%) also demonstrate higher semantic fidelity. This phenomenon profoundly reveals that for SAE explanation generation tasks, utilizing a larger scale of SFT data enables the model to establish a solid mapping from underlying features to natural language. Built upon this foundation, even when reusing the same DPO feature pool across multiple rounds, the model can consistently leverage the closed-loop feedback mechanism to suppress hallucinations and deepen its understanding without suffering from severe overfitting. Conversely, the weaker SFT foundation in exp2 prevents the model from establishing robust initial aligned representations, which subsequently compromises the quality of the constructed preference dataset. Consequently, the preference guidance from new data in later stages fails to compensate for this inherent deficiency in early generative capabilities. Therefore, we primarily adopt the data reuse strategy for all subsequent experiments.

\begin{figure}[htbp]
    \centering
    \includegraphics[width=\linewidth]{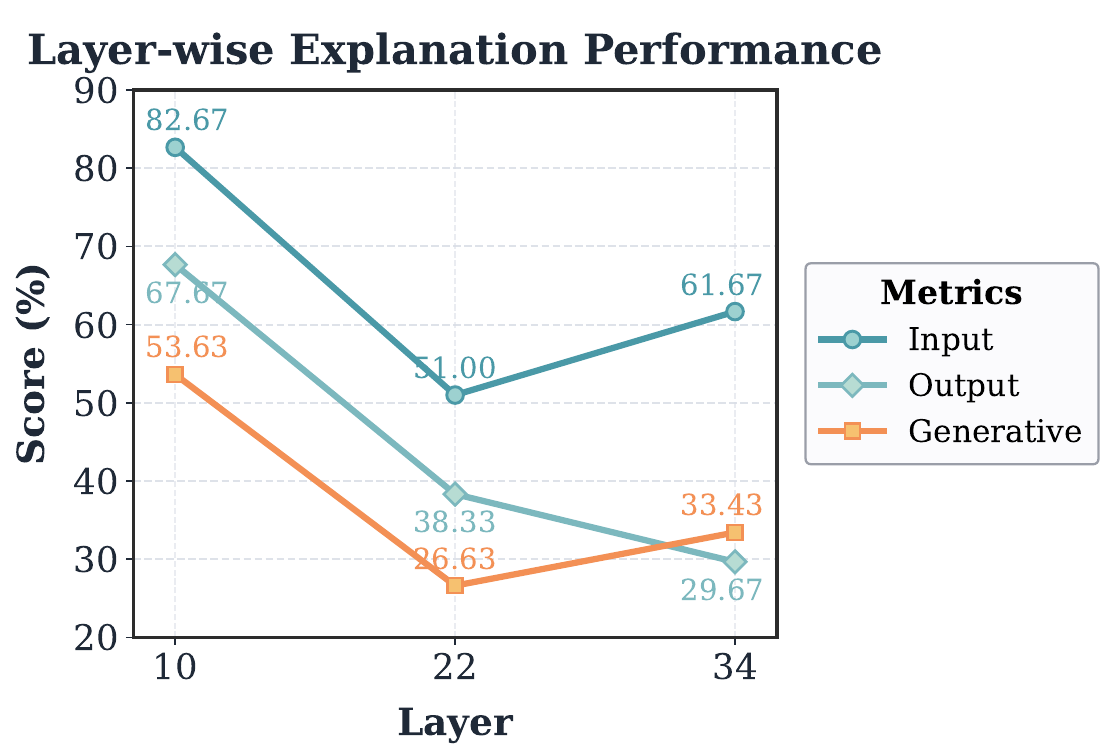}
    \caption{Layer-wise explanation performance evaluated on Gemma-2-27b model.}
    \label{fig:layer_performance_gemma27b}
\end{figure}

\section{Layer-wise Performance Analysis}
To further validate the applicability of SAEExplainer across varying parameter scales and investigate potential scaling behaviors, we conducted a layer-wise performance analysis on the significantly larger Gemma-2-27B model (selecting layers 10, 22, and 34 to represent early, intermediate, and late depths, respectively). 

As illustrated in Figure \ref{fig:layer_performance_gemma27b}, while the 27B model exhibits layer-wise trends similar to those of the 9B model with a pronounced performance dip in the intermediate layer (Layer 22), its quantitative metrics display a much greater magnitude of fluctuation. Despite this increased variance, the overall trajectory aligns with our observations presented in Section \ref{sec:layer_wise_evaluation}, demonstrating the generalizability of SAEExplainer across different model sizes. Regarding the amplified variations observed in the 27B model, we hypothesize that larger models may exhibit fundamentally different feature granularity, activation scales, and internal routing dynamics. We leave a more systematic analysis of these scaling-induced representational differences to future work.

\section{Robustness and Statistical Significance}
\label{app:robustness}

\subsection{Multi-seed Robustness}
\label{app:multi_seed}

To evaluate the robustness of SAEExplainer to training randomness, we conduct multi-seed experiments on Layer 29 of Gemma-2-9B. Specifically, we use four random seeds (42, 43, 44, and 45). For each seed, we randomly sample 5,000 features for SFT training and another 5,000 features for DPO training, and evaluate the resulting model on the same fixed set of 100 held-out features.
As shown in Table~\ref{tab:multi_seed}, SAEExplainer exhibits stable performance across different random seeds, with relatively small standard deviations across all evaluation metrics. These results suggest that the performance of SAEExplainer is robust to random feature sampling and training randomness.

\begin{table}[t]
\centering
\small
\setlength{\tabcolsep}{3.5pt}
\begin{tabular}{@{}lccccc@{}}
\toprule
\textbf{Metric} & \textbf{Set 1} & \textbf{Set 2} & \textbf{Set 3} & \textbf{Set 4} & \textbf{Mean $\pm$ Std} \\
\midrule
GEN & 41.8 & 41.9 & 40.7 & 39.8 & $41.05 \pm 0.99$ \\
INP & 78 & 79 & 75 & 78 & $77.50 \pm 1.73$ \\
OUT & 67 & 70 & 71 & 68 & $69.00 \pm 1.83$ \\
$A_{\mathrm{high}}$ & 45.63 & 47.56 & 44.78 & 45.82 & $45.95 \pm 1.17$ \\
$A_{\mathrm{low}}$ & 0.83 & 1.27 & 0.80 & 0.49 & $0.85 \pm 0.32$ \\
$\Delta A$ & 44.80 & 46.29 & 43.98 & 45.33 & $45.10 \pm 0.97$ \\
\bottomrule
\end{tabular}
\caption{Multi-seed results on Layer 29 of Gemma-2-9B. We report the performance under four random seeds together with the mean and standard deviation.}
\label{tab:multi_seed}
\end{table}

\subsection{Statistical Significance Tests}
\label{app:significance}

To further assess the statistical significance of the improvements achieved by SAEExplainer, we conduct feature-level paired bootstrap tests on Layer 15 of Llama-3.1-8B. For each held-out feature, we compute the score difference between SAEExplainer and the corresponding baseline. We then resample the features with replacement 10,000 times to estimate the 95\% confidence interval (CI) and the $p$-value of the mean improvement.

\begin{table}[t]
\centering
\small
\setlength{\tabcolsep}{3pt}
\renewcommand{\arraystretch}{1.03}
\begin{tabular}{@{}lccccc@{}}
\toprule
\textbf{Metric} & \textbf{Ours} & \textbf{Base.} & \textbf{Diff.} & \textbf{95\% CI} & \textbf{$p$} \\
\midrule
\multicolumn{6}{@{}l}{\textit{Baseline: Neuronpedia}} \\
GEN & 58.4 & 33.9 & +24.5 & [20.2, 28.8] & $<0.0001$ \\
INP & 68.0 & 49.0 & +19.0 & [13.7, 24.3] & $<0.0001$ \\
OUT & 76.0 & 69.3 & +6.7 & [0.7, 12.7] & 0.0358 \\
\addlinespace[2pt]
\multicolumn{6}{@{}l}{\textit{Baseline: GPT-5}} \\
GEN & 58.4 & 52.5 & +5.9 & [2.2, 9.7] & 0.0014 \\
INP & 68.0 & 63.3 & +4.7 & [0.3, 9.0] & 0.0464 \\
OUT & 76.0 & 70.0 & +6.0 & [0.0, 12.0] & 0.0584 \\
\bottomrule
\end{tabular}
\caption{Feature-level paired bootstrap significance tests on Layer 15 of Llama-3.1-8B.}
\label{tab:significance}
\end{table}

As shown in Table~\ref{tab:significance}, SAEExplainer achieves statistically significant improvements over Neuronpedia on all three metrics ($p<0.05$). In particular, SAEExplainer improves generative accuracy and input score by 24.5 and 19.0 points, respectively, while also achieving a significant improvement of 6.7 points on output score.
Compared with the stronger GPT-5 baseline, SAEExplainer also achieves statistically significant improvements on generative accuracy and input score, with $p$-values of 0.0014 and 0.0464, respectively. For output score, SAEExplainer achieves an improvement of 6.0 points, with $p=0.0584$. The improvement remains consistently positive. Overall, these results demonstrate that the performance gains of SAEExplainer are reliable across multiple key evaluation metrics and baseline comparisons.

\section{Human Evaluation}
\label{sec:human_eval}

To further evaluate whether the generated feature explanations accurately characterize the activation conditions of SAE features, we conduct a human evaluation. This evaluation examines whether human annotators can use a given explanation to correctly distinguish texts that truly activate the target feature from semantically similar texts that do not activate it.
Specifically, for each SAE feature, annotators are presented with an anonymized feature explanation and two candidate texts, A and B. One candidate is a true high-activation example, while the other is a negative example with zero activation on the target feature. Based on the provided explanation, annotators are asked to determine which candidate text is more likely to activate the corresponding feature.

To minimize potential annotation bias, we hide the method names, original feature IDs, correct answers, and source question IDs. In addition, the presentation order of features, questions, and A/B candidates is randomized. We recruit three human annotators, each of whom evaluates 50 features, and report the average number of correct judgments for each explanation method.
As shown in Table~\ref{tab:human_eval}, SAEExplainer achieves the highest average human evaluation score of 43.67/50, outperforming all baselines. This indicates that explanations generated by SAEExplainer more effectively help humans identify the true activation conditions of target SAE features and distinguish activating examples from semantically similar non-activating examples. Overall, the human evaluation further demonstrates that SAEExplainer provides more informative explanations for human discrimination, better reflecting the activation patterns associated with the target features.

\begin{table}[t]
\centering
\small
\begin{tabular}{lc}
\toprule
\textbf{Method} & \textbf{Mean Score / 50} \\
\midrule
Activation Oracles & 31.67 \\
Neuronpedia & 31.00 \\
GPT & 37.33 \\
Claude & 40.67 \\
\textbf{SAEExplainer} & \textbf{43.67} \\
\bottomrule
\end{tabular}
\caption{Human evaluation results. The score denotes the average number of correct judgments out of 50 features across three annotators.}
\label{tab:human_eval}
\end{table}

\section{Prompts}\label{app:prompts}

\subsection{SFT Explanation Prompt}
\label{prompt sft}

\begin{tcolorbox}[
    enhanced,
    breakable,
    colback=orange!10!white, colframe=blue!5!black,
    arc=2mm,
    boxrule=1pt,
    title={\bfseries SFT Explanation Prompt},
    coltitle=white,
    attach boxed title to top left={yshift=-2mm, xshift=3mm},
    boxed title style={enhanced, colback=blue!5!black, colframe=blue!5!black, arc=2mm, boxrule=0pt},
    top=0.5mm, left=1mm, right=1mm, bottom=0.5mm,
]
\small\vskip8pt
The internal neural feature represented by \texttt{\{placeholder\_text\}} is provided only through internal neural conditioning. Using the injected feature representation, provide one clear, accurate explanation of the feature.
\end{tcolorbox}

\subsection{Preference Generation Prompt}
\label{prompt generation}

\begin{tcolorbox}[
    enhanced,
    breakable,
    colback=orange!10!white, colframe=blue!5!black,
    arc=2mm,
    boxrule=1pt,
    title={\bfseries Preference Generation Prompt},
    coltitle=white,
    attach boxed title to top left={yshift=-2mm, xshift=3mm},
    boxed title style={enhanced, colback=blue!5!black, colframe=blue!5!black, arc=2mm, boxrule=0pt},
    top=0.5mm, left=1mm, right=1mm, bottom=0.5mm,
]
\small\vskip8pt
You are an expert corpus-construction specialist and careful semantic writer. Your task is to write short texts that faithfully and unmistakably express one given semantic concept.

Please carefully read the following [Core Concept]. Treat it as the exact semantic target, not a loose theme:

\texttt{[\{explanation\}]}

Based on this concept, create \texttt{\{num\_samples\}} completely independent short texts.

\textbf{[Strict Requirements]}
\begin{itemize}[nosep, topsep=2pt, itemsep=0pt, parsep=0pt, leftmargin=0.7cm]
    \item Semantic fidelity is the top priority: every text must clearly express the exact core concept above. Do not broaden it into neighboring concepts, adjacent domains, or vaguely related themes.
    \item Stay on-concept even when varying the surface form: diversity should come from wording, scenario details, or discourse style, not from changing the underlying semantic meaning.
    \item Hidden and natural (show, don't tell): do not define the concept explicitly as if writing a dictionary. Do not say things like ``this text illustrates...''. Instead, embed the concept naturally into a concrete situation.
    \item Extremely concrete: use specific scenes, actions, terminology, or dialogue so that the concept is expressed with high semantic density.
    \item Do not introduce strong domain cues unless they are genuinely implied by the concept itself.
    \item Avoid generic topical similarity: a text is bad if it merely feels loosely related while failing to make a careful reader recover the same concept.
    \item Short and compact: each text must be between 15 and 30 words.
    \item Moderate diversity only: the \texttt{\{num\_samples\}} texts should not be duplicates, but they should remain tightly centered on the same concept.
\end{itemize}

Before answering, silently check each text: would an independent reader likely summarize it with the same core concept above? If not, rewrite it.

You must return only valid JSON with a top-level field named \texttt{"samples"}. Do not include Markdown code fences or any extra explanation.

\texttt{\{}
\texttt{\ \ "samples": [}
\texttt{\ \ \ \ "Text 1...",}
\texttt{\ \ \ \ "Text 2...",}
\texttt{\ \ \ \ "..."}
\texttt{\ \ ]}
\texttt{\}}
\end{tcolorbox}

\subsection{Generative Evaluation Prompts}
\label{Prompt generative}

\begin{tcolorbox}[
    enhanced,
    breakable,
    colback=orange!10!white, colframe=blue!5!black,
    arc=2mm,
    boxrule=1pt,
    title={\bfseries System Prompt},
    coltitle=white,
    attach boxed title to top left={yshift=-2mm, xshift=3mm},
    boxed title style={enhanced, colback=blue!5!black, colframe=blue!5!black, arc=2mm, boxrule=0pt},
    top=0.5mm, left=1mm, right=1mm, bottom=0.5mm,
]
\small\vskip8pt
You are an expert test case generator for Sparse Autoencoder features. You generate diverse, high-quality texts that strongly activate one feature.
\end{tcolorbox}

\begin{tcolorbox}[
    enhanced,
    breakable,
    colback=orange!10!white, colframe=blue!5!black,
    arc=2mm,
    boxrule=1pt,
    title={\bfseries User Prompt},
    coltitle=white,
    attach boxed title to top left={yshift=-2mm, xshift=3mm},
    boxed title style={enhanced, colback=blue!5!black, colframe=blue!5!black, arc=2mm, boxrule=0pt},
    top=0.5mm, left=1mm, right=1mm, bottom=0.5mm,
]
\small\vskip8pt
You are an expert test case generator for Sparse Autoencoder (SAE) features. Your task is to generate diverse test sentences that strongly activate the given feature.

\textbf{FEATURE DESCRIPTION:}

\texttt{\{explanation\}}

\textbf{TASK:}

Generate exactly \texttt{\{num\_samples\}} diverse test sentences that should activate this feature HIGHLY. Each sentence must clearly demonstrate the pattern or behavior described in the feature description.

\textbf{REQUIREMENTS:}
\begin{itemize}[nosep, topsep=2pt, itemsep=0pt, parsep=0pt, leftmargin=0.7cm]
    \item Sentence length: 15--30 words each
    \item Diversity: Vary contexts, phrasings, subjects, and scenarios
    \item Clarity: Each sentence should unambiguously match the feature description
    \item Naturalness: Use natural, grammatically correct English
    \item Uniqueness: Avoid repetitive or overly similar sentences
    \item Coverage: Test different aspects and variations of the described pattern
\end{itemize}

\textbf{OUTPUT FORMAT:}

Return only valid JSON with a top-level field named \texttt{"samples"}.

Do not include Markdown code fences or any extra explanation.

\texttt{\{}
\texttt{\ \ "samples": [}
\texttt{\ \ \ \ "Sentence 1...",}
\texttt{\ \ \ \ "Sentence 2...",}
\texttt{\ \ \ \ "..."}
\texttt{\ \ ]}
\texttt{\}}

\textbf{IMPORTANT:}
\begin{itemize}[nosep, topsep=2pt, itemsep=0pt, parsep=0pt, leftmargin=0.7cm]
    \item Output ONLY the JSON object, with no additional commentary
    \item Do not include quotes around the whole response
    \item Ensure each sentence is distinct and tests different variations of the pattern
    \item Focus on sentences that would produce strong activation values for this feature
\end{itemize}
\end{tcolorbox}

\subsection{Input Metric Prompts}
\label{prompt input}

\begin{tcolorbox}[
    enhanced,
    breakable,
    colback=orange!10!white, colframe=blue!5!black,
    arc=2mm,
    boxrule=1pt,
    title={\bfseries Pos/Neg Examples Generation Prompt},
    coltitle=white,
    attach boxed title to top left={yshift=-2mm, xshift=3mm},
    boxed title style={enhanced, colback=blue!5!black, colframe=blue!5!black, arc=2mm, boxrule=0pt},
    top=0.5mm, left=1mm, right=1mm, bottom=0.5mm,
]
\small\vskip8pt
I'm going to give you explanations and interpretations of features from LLMs. You must take in each explanation, and generate 5 sentences for which you think the feature will have a high activation, and 5 for which they'll have a low activation. For the high activation, make sure to choose ones that will cause a high activation with high confidence -- you don't have to include all groups, just make examples that you're confident will have high activation. Make the sentences both include the words from the explanation, and represent the concept. Try to use specific examples, and make them literal interpretations of the explanation, without trying to generalize. Low activation sentences should have nothing to do with the interpretation, i.e. they should be orthogonal and completely unrelated. Please output the response in JSON format with a \texttt{"positive"} key and a \texttt{"negative"} key. Output only the JSON and no other explanation. Make sure the JSON is formatted correctly -- do not include any backticks, and do not format as code. The explanations should be five and five overall, not per line.

\texttt{\{explanation\}}
\end{tcolorbox}

\subsection{Output Metric Prompts}
\label{prompt output}

\begin{tcolorbox}[
    enhanced,
    breakable,
    colback=orange!10!white, colframe=blue!5!black,
    arc=2mm,
    boxrule=1pt,
    title={\bfseries Fixed Generation Prompt},
    coltitle=white,
    attach boxed title to top left={yshift=-2mm, xshift=3mm},
    boxed title style={enhanced, colback=blue!5!black, colframe=blue!5!black, arc=2mm, boxrule=0pt},
    top=0.5mm, left=1mm, right=1mm, bottom=0.5mm,
]
\small\vskip8pt
The three fixed prompts used to generate steered output bundles are:
\begin{itemize}
    \item \texttt{The explanation is simple:}
    \item \texttt{I think}
    \item \texttt{We}
\end{itemize}
\end{tcolorbox}

\begin{tcolorbox}[
    enhanced,
    breakable,
    colback=orange!10!white, colframe=blue!5!black,
    arc=2mm,
    boxrule=1pt,
    title={\bfseries Judge Prompt},
    coltitle=white,
    attach boxed title to top left={yshift=-2mm, xshift=3mm},
    boxed title style={enhanced, colback=blue!5!black, colframe=blue!5!black, arc=2mm, boxrule=0pt},
    top=0.5mm, left=1mm, right=1mm, bottom=0.5mm,
]
\small\vskip8pt
You are analyzing the behavior of a specific neuron in a language model. You will receive:

\begin{enumerate}[nosep, topsep=2pt, itemsep=0pt, parsep=0pt, leftmargin=0.8cm]
    \item A hypothesized explanation for what concept the neuron represents (e.g., specific tokens, themes, or ideas).
    \item Three sets of completions, one generated by amplifying the activation of the neuron in question, and two generated by amplifying two different random neurons across the same prompts.
\end{enumerate}

Your goal is to identify which set of completions is more likely the result of amplifying the neuron in question. To do this:
\begin{itemize}[nosep, topsep=2pt, itemsep=0pt, parsep=0pt, leftmargin=0.7cm]
    \item Look for completions where the \textbf{literal words} or the \textbf{ideas/themes} described in the explanation occur more frequently or with greater emphasis.
    \item Remember that amplification may highlight specific words or their broader contextual meanings, meaning that many outputs may be noisy, but still contain keywords that appear in the explanation.
    \item Your answer should be based on the \textbf{content} of the completions, not the quality of the language model's output.
    \item Your reasoning should be sound; do not make overly elaborate or far-fetched connections.
\end{itemize}

The first line in your response should be a brief explanation of your choice -- what made you choose that set of completions.

The second line must be only the set number you think matches the description (i.e., 1, 2 or 3) and no other text. You must pick one of the three sets.

\vspace{3mm}

\texttt{Explanation: \{explanation\}}

\vspace{2mm}
\texttt{\# Set 1}

\texttt{\{amplifications1\}}

\vspace{2mm}
\texttt{\# Set 2}

\texttt{\{amplifications2\}}

\vspace{2mm}
\texttt{\# Set 3}

\texttt{\{amplifications3\}}
\end{tcolorbox}

\section{Examples of SAEExplainer Explanations}
Table \ref{tab:comprehensive_sae_features} provides a comprehensive selection of qualitative examples to illustrate the feature explanations generated by SAEExplainer. 
\begin{table*}[t!] 
\centering
\small
\renewcommand{\arraystretch}{1.4} 

\begin{tabularx}{\textwidth}{@{} >{\raggedright\arraybackslash}p{6.5cm} >{\raggedright\arraybackslash}X @{}}
\toprule
\textbf{Feature Path} & \textbf{Explanation} \\ \midrule

\rowcolor{gray!10} \multicolumn{2}{@{}l}{\textbf{Model: Gemma-2-9b}} \\ \addlinespace
\texttt{gemma-2-9b/12-gemmascope-res-16k/146}   & terms related to inflation and its effects on the economy, particularly focusing on inflation rates and monetary policy responses \\ \addlinespace
\texttt{gemma-2-9b/12-gemmascope-res-16k/10914} & statistical terms and confidence intervals related to estimations in data analysis \\ \addlinespace
\texttt{gemma-2-9b/12-gemmascope-res-16k/9338}  & PHP code related to password hashing and verification using the bcrypt algorithm \\ \addlinespace
\texttt{gemma-2-9b/31-gemmascope-res-16k/15217} & technical terms related to Bluetooth Low Energy (BLE) and its specifications \\ \addlinespace
\texttt{gemma-2-9b/37-gemmascope-res-16k/11808} & scientific terminology related to nuclear physics and the nuclear shell model \\ \addlinespace

\rowcolor{gray!10} \multicolumn{2}{@{}l}{\textbf{Model: Gemma-2-27b}} \\ \addlinespace
\texttt{gemma-2-27b/10-gemmascope-res-131k/75318}  & The neuron fires on the discourse marker ``Further,'' (as in ``Further, we show\dots'') that introduces an additional point in academic or technical writing. \\ \addlinespace
\texttt{gemma-2-27b/10-gemmascope-res-131k/127747} & The neuron fires on words and phrases related to taming or domesticating---e.g., ``tame,'' ``taming,'' ``tamed,'' ``domesticate,'' ``The Taming of\dots,'' ``Wild West tamed,'' etc. \\ \addlinespace
\texttt{gemma-2-27b/22-gemmascope-res-131k/17915}  & The neuron fires on mentions of Wi-Fi and related wireless-network configuration terms (e.g., ``wifi,'' ``ssid,'' ``wireless,'' ``network,'' ``WPA,'' ``802.11,'' etc.). \\ \addlinespace
\texttt{gemma-2-27b/22-gemmascope-res-131k/67616}  & The neuron fires on reassuring medical/health statements that a condition is normal or benign (e.g., ``this is normal and nothing to worry about''). \\ \addlinespace
\texttt{gemma-2-27b/34-gemmascope-res-131k/49161}  & The neuron fires on Java servlet signature tokens---especially the \texttt{HttpServletRequest}, \texttt{HttpServletResponse}, and related parameter types (e.g., ``request'', ``response'') in method declarations. \\ \addlinespace
\texttt{gemma-2-27b/34-gemmascope-res-131k/9393}   & The neuron fires on mentions of ``sense of humor'' (and its variants like ``deadpan,'' ``dry,'' ``sarcastic,'' etc.), i.e., phrases describing someone's humorous sensibility. \\ \addlinespace

\rowcolor{gray!10} \multicolumn{2}{@{}l}{\textbf{Model: Llama-3.1-8b}} \\ \addlinespace
\texttt{llama3.1-8b/15-llamascope-res-32k/12980} & \LaTeX~document class and package declarations \\ \addlinespace
\texttt{llama3.1-8b/15-llamascope-res-32k/5231}  & references to design elements and the visibility of materials or construction in architecture and fashion \\ \addlinespace
\texttt{llama3.1-8b/15-llamascope-res-32k/22068} & statistics related to soccer match possession and chances, particularly focusing on half-time and overall match dynamics \\ \addlinespace
\texttt{llama3.1-8b/25-llamascope-res-32k/10353} & instructions related to artistic techniques, particularly those involving layering and coloring with mediums like ink and watercolors \\ \addlinespace
\texttt{llama3.1-8b/25-llamascope-res-32k/932}   & descriptions of garment features, particularly those related to sleeves and neckline details in sewing patterns \\ \addlinespace
\texttt{llama3.1-8b/25-llamascope-res-32k/623}   & information about musical lineups and performances, particularly focusing on headliners and supporting acts in festivals or concerts \\ \bottomrule
\end{tabularx}
\caption{Comprehensive qualitative examples of learned SAE features across different base models (Gemma-2-9b, Gemma-2-27b, and Llama-3.1-8b), illustrating the diversity and specificity of the concepts captured.}
\label{tab:comprehensive_sae_features}
\end{table*}

\end{document}